\pdfoutput=1
\documentclass[11pt]{article}

\usepackage[table]{xcolor}
\usepackage[]{acl}

\usepackage{times}
\usepackage{latexsym}

\usepackage[T1]{fontenc}
\usepackage[utf8]{inputenc}

\usepackage{microtype}

\usepackage{color}
\usepackage{arydshln}
\usepackage{booktabs}
\usepackage{graphicx}
\usepackage{subcaption}
\usepackage{float}
\usepackage{amsmath,amsfonts,amssymb,bbm,epsfig,bm}
\usepackage[ruled,noresetcount,linesnumbered]{algorithm2e}
\usepackage{listings}
\usepackage{url}
\usepackage{spverbatim}
\usepackage{multirow}
\usepackage{pbox}
\usepackage{tikz}
\usepackage{pifont}
\usepackage{xspace}
\usepackage{physics}
\usepackage[nameinlink,capitalize]{cleveref}
\usepackage{scalerel}
\usepackage[font=small]{caption}
\usepackage{titlesec}
\usepackage{todonotes}
\usepackage[normalem]{ulem}
\usepackage{sidecap}
\usepackage{comment}

\DeclareSymbolFont{extraup}{U}{zavm}{m}{n}
\DeclareMathSymbol{\varheart}{\mathalpha}{extraup}{86}
\DeclareMathSymbol{\vardiamond}{\mathalpha}{extraup}{87}

\makeatletter
\Crefname{algorithm}{Algo.}{Algorithms}
\Crefname{table}{Tab.}{Tables}
\crefname{section}{\S\@gobble}{\S\S\@gobble}
\crefname{subsection}{\S\@gobble}{\S\S\@gobble}
\makeatother

\newcommand{\vcenteredinclude}[1]{\begingroup
\setbox0=\hbox{\includegraphics[height=1.0em]{#1}}%
\parbox{\wd0}{\box0}\endgroup}

\titlespacing{\paragraph}{%
  0pt}{%              left margin
  0.3\baselineskip}{% space before (vertical)
  1em}%               space after (horizontal)

\renewcommand{\tt}[1]{\fontfamily{cmtt}\selectfont #1}

\definecolor{darkgreen}{RGB}{43,163,39}
\definecolor{amaranth}{rgb}{0.9, 0.17, 0.31}

\newcommand{\yin}[1]{{\color{red} [Pengcheng: {#1}]}}

\newcommand{\eg}{\hbox{\emph{e.g.}}\xspace}

\definecolor{deepblue}{rgb}{0,0,0.5}
\definecolor{deepred}{rgb}{0.6,0,0}
\definecolor{deepgreen}{rgb}{0,0.5,0}
\definecolor{darkgreen}{RGB}{43,163,39}
\definecolor{bluesquare}{rgb}{126,166,224}

\newcommand{\xmark}{\vcenteredinclude{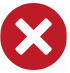}}
\newcommand{\cmark}{\vcenteredinclude{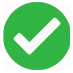}}
\newcommand{\filterfn}{\vcenteredinclude{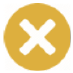}}
\newcommand{\filterfp}{\vcenteredinclude{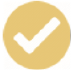}}

\newcommand{\hbpaper}{HB19}

\def\overnight/{\textsc{OverNight}}
\def\granno/{\textsc{Granno}}
\def\scholar/{\textsc{Scholar}}
\def\geo/{\textsc{Geo}}

\newcommand\utterance[1]{\textit{#1}}
\newcommand\mr{\ensuremath{\bm{z}}}
\newcommand\utt{\ensuremath{\bm{u}}}

\newcommand\uttnat{\ensuremath{\bm{u}_\textrm{nat}}}

\newcommand{\Dnat}{\ensuremath{\mathbb{D}_\textrm{nat}}}
\newcommand{\Dpara}{\ensuremath{\mathbb{D}_\textrm{par}}}
\newcommand{\Dcan}{\ensuremath{\mathbb{D}_\textrm{can}}}
\newcommand{\Dfcan}{\ensuremath{\mathbb{D}_\textrm{can}}}

\newcommand\std[1]{\ensuremath{\scriptstyle \pm #1}}
\newcommand\notenum[1]{\ensuremath{\scriptstyle (#1)}}

\lstdefinestyle{program}{
  basicstyle=\fontfamily{cmtt}\small,
  language=Python,
  otherkeywords={self,call,append,split,write},             % Add keywords here
  keywordstyle=\bfseries\color{deepblue},
  emph={},          % Custom highlighting
  emphstyle=\color{deepred},    % Custom highlighting style
  showstringspaces=false,
  breaklines=true,
  escapeinside=||,
  columns=fullflexible,
}

\title{On The Ingredients of an Effective Zero-shot Semantic Parser}

\author{
	Pengcheng Yin$^\spadesuit$ \quad John Wieting$^\clubsuit$ \quad Avi Sil$^\vardiamond$ \quad Graham Neubig$^\spadesuit$ \\ 
	$^\spadesuit$Carnegie Mellon University \quad $^\clubsuit$Google Research \quad $^\vardiamond$IBM Research \\
	{\tt \{pcyin,gneubig\}@cs.cmu.edu} \quad {\tt jwieting@google.com} \quad {\tt avi@us.ibm.com}
}

\begin{document}
\maketitle
\begin{abstract}
%\gn{Title candidate: ``On The Ingredients of an Effective Zero-shot Semantic Parser'', it sounds a little bit more interesting than just ``Improving''. Alternatively, you could be a little bit more concrete about the individual contributions? Also, I think ``Neural'' probably isn't necessary.}
Semantic parsers map natural language utterances into meaning representations (\eg~programs).
%\gn{Not sure this is exactly accurate, what about AMR parsers for example?}.
%Recently, there has been a burgeoning in developing zero-shot neural semantic parsers without manually curated training data \gn{In general, I find that it's often better to replace these ``recently'' sentences with something more fundamental about the problem. For example, explain that annotation of semantic parses is laborious, and zero-shot is a way around it?}.
%However, training data for such parsers is often limited due to the difficulty in finding annotators fluent in the language in which such programs are expressed.
Such models are typically bottlenecked by the paucity of training data due to the required laborious annotation efforts.
%Thus, recent studies have performed zero-shot learning by automatically synthesizing training examples of canonical utterances and programs from a synchronous grammar, and further paraphrasing these utterances using pre-trained language models (LMs) to improve their naturalness and linguistic diversity.
Recent studies have performed zero-shot learning by  synthesizing training examples of canonical utterances and programs from a grammar, and further paraphrasing these utterances to improve linguistic diversity.
%better match with real-world  utterances.
%In this paper, we aim to improve the performance of existing zero-shot neural semantic parsers, while trying to understand the key factors that contribute to their performance \gn{Maybe talk about understanding first, and then improving after that? It seems that the understanding would lead you to a model design that allows you to make improvements.}.
However, such synthetic examples cannot fully capture patterns in real data. 
%such models are subject to the fundamental mismatch between the synthetic and real-world user-issued examples.
In this paper we analyze zero-shot parsers through the lenses of the \emph{language} and \emph{logical} gaps \cite{herzig19dontdetect}, which quantify the discrepancy of language and programmatic patterns between the synthetic canonical examples and real-world user-issued ones.
We propose bridging these gaps using improved grammars, stronger paraphrasers, and efficient learning methods using canonical examples that most likely reflect real user intents.
Our model achieves strong performance on two semantic parsing benchmarks (\textsc{Scholar}, \textsc{Geo}) with \emph{zero} labeled data.

%In this paper we aim to understand the key factors that contribute to the performance of such zero-short semantic parsers, and use this understanding to improve their performance.
%Specifically, we analyze existing zero-shot parsers through the lenses of the \emph{language} and \emph{logical} gaps \cite{herzig19dontdetect}, which quantify the discrepancy of language and programmatic patterns between the machine-synthesized canonical examples and real-world user-issued ones.
% Our analysis suggests that while language gaps could be largely reduced using carefully designed grammars and stronger paraphrasers based on pre-trained LMs, logical gaps, due to the limited coverage of the grammar and the quality of the paraphraser, still remain a bottleneck.
% From these insights, we propose methods to bridge these gaps with improved grammars and paraphrasers, which achieve strong performance on two semantic parsing benchmarks (\textsc{Scholar}, \textsc{Geo}) with \emph{zero} labeled data.
%\gn{what is the methodology you propose to achieve strong performance?}.
%\yin{Revise the abstract}
\end{abstract}

\section{Introduction}

Semantic parsers translate natural language (NL) utterances into formal meaning representations.
In particular, task-oriented semantic parsers map user-issued utterances (\eg~\textit{Find papers in ACL}) into machine-executable programs (\eg~a database query), 
%Such models provide natural language interfaces to computational systems, and 
play a key role in providing natural language interfaces to applications like conversational virtual assistants~\citep{gupta18task,machines2020task-oriented}, 
% \gn{``Semantic Machines et al.'' is a bit strange, as that means ``Semantic Machines and others'' but there are not any people unaffiliated with Semantic Machines who wrote this paper...}\yin{Yeah, but that's the official bibtex entry. Is there any way to suppress et al for a single bibtex entry?}\gn{I don't see it in the official bibtex entry? \url{https://www.aclweb.org/anthology/2020.tacl-1.36.bib}}
robot instruction following~\citep{artzi-zettlemoyer:2013:TACL,fried2018speaker}, as well as querying databases~\citep{Li2014ConstructingAI,Yu2018SpiderAL} or generating Python code~\citep{yin17acl}.
%using natural language~\citep{yin17acl}\yin{More cite here}.

%Research in this area could be broadly categorized into two lines.
%The first line of work considers 
%There are two lines of works in semantic parsing. The first line concerns with representing meanings of sentences using formal structures like AMR. 
%The second line, which is the focus of this paper, aims to develop natural language understanding systems that map user-issued utterances into executable programs. Such systems play a key role in tasks like question answering and conversational AI.
%Recent years have witnessed significant progress in building such systems (cite), and advance in this line enables commerial products like Siri and Alexa.

%\paragraph{TODO: Difficulty in Collecting Parallel Data} However, a key issue in semantic parsing is collecting annotated data...

%Learning semantic parsers typically requires parallel data of NL utterances annotated with programs.
%However, annotating such data requires significant expertise and cost~\citep{berant2013freebase},
Learning semantic parsers typically requires parallel data of utterances annotated with programs, which requires significant expertise and cost~\citep{berant2013freebase}.
Thus, the field has explored alternative approaches using supervisions cheaper to acquire, such as the execution results~\citep{DBLP:conf/conll/ClarkeGCR10} or unlabeled utterances~\citep{poon13grounded}.
%or semi-supervised learning, which uses extra unlabeled NL utterances in addition to the limited parallel labeled data.
In particular, the seminal \overnight/ approach~\citep{wang15overnight} synthesizes parallel data by using a synchronous grammar to align programs and their canonical NL expressions 
(\eg~{\tt Filter(paper,venue=\fbox{?})} $\leftrightarrow$ \textit{papers in \fbox{?}} and {\tt acl}$\leftrightarrow$\textit{ACL}),
%(\eg~{\tt CreateEvent()} $\leftrightarrow$ \textit{Create an event} and {\tt attendees=Jane} $\leftrightarrow$ \textit{with Jane}), 
then generating examples of compositional utterances (\eg \textit{Papers in ACL}) with programs (\eg~{\tt Filter(paper,venue=acl)}). The synthesized utterances are paraphrased by annotators, a much easier task than writing programs.

\begin{figure*}[t]
  \centering{\phantomsubcaption\label{fig:system:grammar}\phantomsubcaption\label{fig:system:candata}\phantomsubcaption\label{fig:system:iter}}
  \includegraphics[width=1. \textwidth]{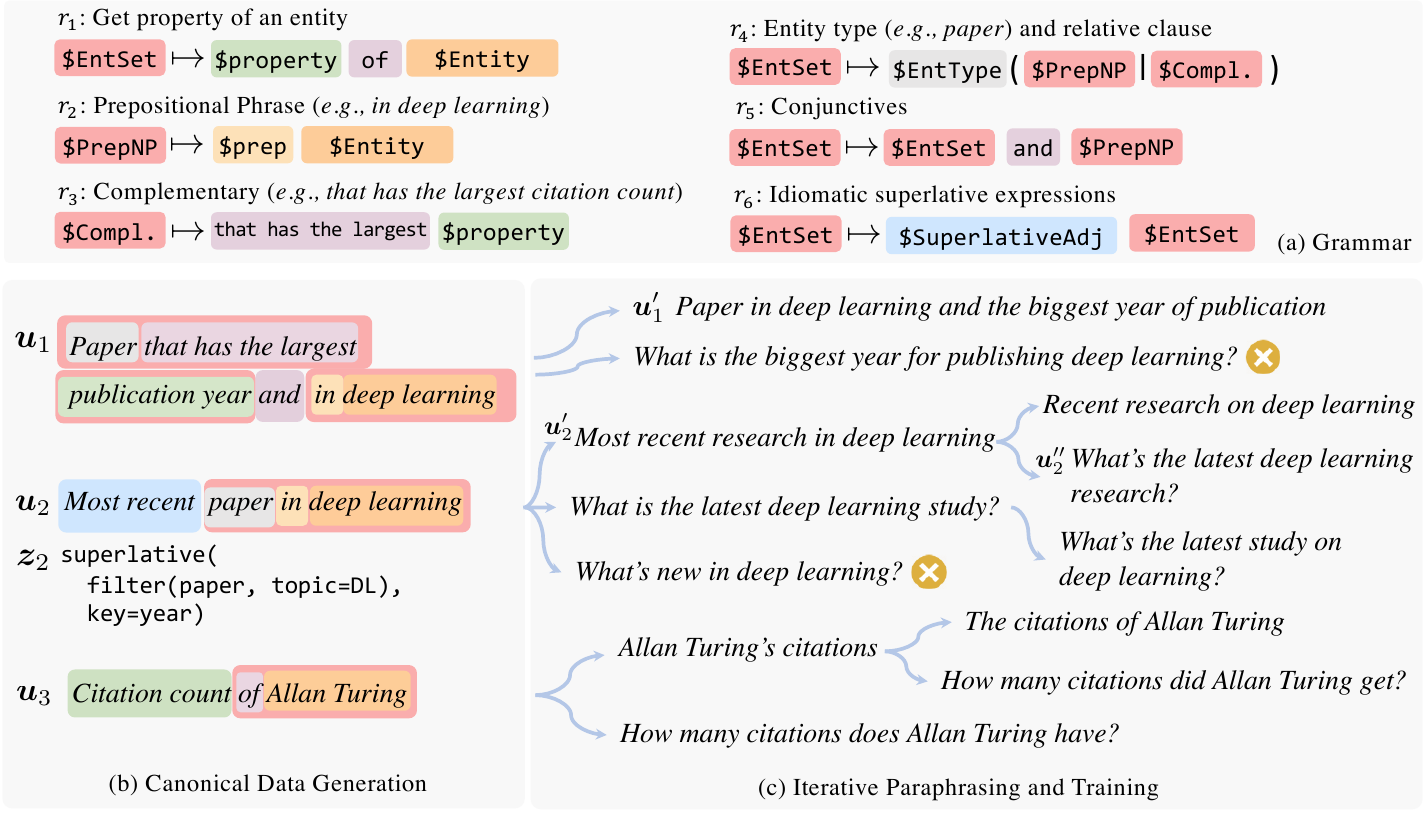}
  \caption{Illustration of the learning process of our zero-shot semantic parser with real model outputs. \textbf{(a)} Synchronous grammar with production rules. \textbf{(b)} Canonical examples of utterances with programs (only $\mr_2$ is shown) are generated  from the grammar (colored spans show productions used). Unnatural utterances like $\utt_1$ can be discarded, as in \cref{sec:bridge_gaps:data_selection} \textbf{(c)} At each iteration, canonical examples are paraphrased to increase diversity in language style, and a semantic parser is trained on the paraphrased examples. Potentially noisy or vague paraphrases are filtered (marked as \filterfn) using the parser trained on previous iterations. 
%   \jw{One thing that I stumbled on when looking at this, is u1 doesn't make sense to me. Maybe some explanation about how this is valid from the grammar but should be filtered out early as its not a good thing to be spending training time on (i.e. the naturalness filter you propose)? Also the rejected paraphrases seem plausible to me - {\it noisy} makes it seem like they are wrong, but sometimes they are just too vague or filtered by mistake. I think this should be pointed out here too if possible to make it easier to understand how your model works.}\yin{I edited the text a bit, not sure if it's good...}
  }
  \label{fig:system}
  \vspace*{-5mm}
\end{figure*}

Recently, \citet{Xu2020AutoQA} build upon \overnight/ and develop a \emph{zero-shot} semantic parser replacing the manual paraphrasing process with an automatic paraphrase generator (\cref{sec:system}).
%\paragraph{TODO: Introduce Zero-shot Parsing} Zero-shot parsing has become possible with pre-trained LMs as powerful paraphrasers. Introduce AutoQA and the underlying OverNight procedure (canonical grammar $\mapsto$ canonical examples $\mapsto$ human or automate paraphrasing $\mapsto$ final data).
%While methods like \overnight/ significantly mitigates the amount of data annotation efforts, it still requires manual paraphrasing.
%\yin{TODO: Ann one sentence describing Turker paraphrasing is hard.}
%To enable learning parsers without using labeled training data, xxx recently propose a \emph{zero-shot} semantic parsing approach that leverages off-the-shelf paraphrase generation models to automatically paraphrase machine-synthesized canonical utterances. 
%While promising, there are still several issues with the current approach.
%While promising, it still has several issues.
While promising, there are still several open challenges.
First, such systems are not truly zero-shot --- they still require labeled validation data (\eg~to select the best checkpoint at training).
Next, to ensure the quality and broad-coverage of synthetic canonical examples, those models rely on heavily curated grammars (\eg~with 800 production rules), which are cumbersome to maintain.
%Additionally, to ensure the naturalness and broad-coverage of synthesized canonical utterances, existing models rely on a heavily curated grammar with 800 production rules~\citep{xucikmpaper}, which is cumbersome to maintain.
%Generation from such a large grammar could yield tens of thousands of canonical samples, and paraphrasing and training the parser on all of them could be computationally expensive.
More importantly, as suggested by \citet{herzig19dontdetect} who study \overnight/ models using manual paraphrases, 
%\gn{This paper is not in the context of zero-shot parsing, right? I think it might be important to mention this, as one major contribution of our paper is application of these methods to zero-shot parsing.}, 
such systems trained on synthetic samples suffer from fundamental mismatches between the distributions of the automatically generated examples and the \emph{natural} ones issued by real users. 
Specifically, there are two types of gaps.
%First, there is a \emph{logical gap} between the synthetic and real programs, as real utterances may exhibit logic patterns outside of the domain of those covered by the grammar.
First, there is a \emph{logical gap} between the synthetic and real programs, as real utterances (\eg~\utterance{Paper coauthored by Peter and Jane}) may exhibit logic patterns outside of the domain of those covered by the grammar (\eg~\utterance{Paper by Jane}).
The second is the \emph{language gap} between the synthetic and real utterances, as paraphrased utterances (\eg~$\utt'_1$ in \cref{fig:system}) still follow similar linguistic patterns as the canonical ones they are paraphrased from (\eg~$\utt_1$), while user-issued utterances are more linguistically diverse (\eg~$\utt_2$). 
In this paper we analyze zero-shot parsers through the lenses of language and logical gaps, and propose methods to close those gaps~(\cref{sec:gaps}).
%which quantify the discrepancy of language and programmatic patterns between the machine-synthesized canonical examples and real-world user-issued ones.
%We find that there are significant gaps in these two dimensions for vanilla models.
%We observe vanilla models suffer from such gaps, with generated 
%As we explain in \cref{sec:exp}, a language model fine-tuned on canonical data attained high perplexity on natural user-generated utterances, indicating high language gap.
%On the other hand, logical gaps are also inevitable.
%This is because canonical examples are generated by exhaustively enumerating all possible programs from the synchronous grammar up to a certain depth of programs, and increasing the maximum generation depth to cover more complex real examples will eventually lead to exponentially more canonical samples, which is computationally intractable.
%From such insights, in this paper we study methods to bridge the language and logical gaps for zero-shot semantic parsing.
Specifically, we attempt to bridge the language gap using stronger paraphrasers and more expressive grammars tailored to the domain-specific idiomatic language patterns.
% Instead of hand-crafting a large set of grammatical rules 
%to cover diverse linguistic patterns 
% which could be difficult to maintain, we use a 
We replace the large grammars of previous work with a highly compact grammar % \jw{Could say here how this compares to grammars of related papers to show how different this is.}
with only 46 domain-general production rules, plus a small set of domain-specific productions to capture idiomatic language patterns (\eg~$\utt_2$ in \cref{fig:system}, \cref{sec:bridge_gaps:idiomatic_rules}).
 %(\eg~\textit{``\underline{Most popular} dataset''}).
%that commonly occur in real utterances of the domain.
%Specifically, we improve the language coverage of canonical examples by designing production rules to capture domain-specific language patterns commonly occur in real utterances.
%This could be special rules to form compositional queries as compound nouns (\eg~\textit{\underline{ACL} \underline{2017} \underline{semantic parsing} papers}), or succinct, non-compositional phrases for multi-hop relations (\eg~\textit{Researchers \underline{in} deep learning} v.s.~\textit{Researcher that \underline{write} paper whose \underline{topic} is deep learning}).
%or task-specific expressions for superlative queries (\eg~\textit{\underline{Most productive} author}).
We demonstrate that models equipped with such a smaller but more expressive grammar catered to the domain 
%and state-of-the-art paraphrasers 
could generate utterances with more idiomatic and diverse language styles.
%yielding achieve strong results on benchmarks featuring real-world utterances.

On the other hand, closing the logical gap is non-trivial, since canonical examples are generated by exhaustively enumerating all possible programs from the grammar up to a certain depth, and increasing the threshold to cover more complex real-world examples will lead to exponentially more canonical samples, the usage of which is computationally intractable.
%Additionally, to tackle the exponentially exploding space of canonical examples in order to cover complex programs to bridge the logical gap, we propose a data-efficient approach to sample canonical examples from the grammar as training data.
To tackle the exponentially exploding sample space, we propose an efficient sampling approach by 
%close the logical gap, we propose an efficient approach to sample canonical examples from the synchronous grammar.
%while increasing the exponentially exploding search space to cover more complex examples.
%To increase the generation depth to cover examples with more complex programs while maintaining tractability in the exponentially exploding search space, 
retaining canonical samples that most likely appear in real data (\cref{sec:bridge_gaps:data_selection}).
%Specifically, we approximate the ``naturallness'' of synthesized canonical examples using  probabilities of their utterances measured by pre-trained language models (LMs).
Specifically, we approximate the likelihood of canonical examples using the probabilities of their utterances measured by pre-trained language models (LMs).
This enables us to improve logical coverage of programs 
%with increased generation depth 
while maintaining a tractable number of highly-probable examples as training data.
%\gn{We haven't started talking about experiments yet, so it seems a little strange to have the following two sentences here. Maybe they could be moved to the part where we talk about experiments, or perhaps removed altogether.}
%More surprisingly, our experiments suggest that including more canonical examples in the training data could actually be counter-productive, as most canonical samples are not \emph{useful} since they are unlikely to reflect the intents of real users.
%Our data-efficient approach, on the other hand, could effectively train a parser using only 1\% highly-likely examples from the exponential space.

In experiments, we show that by bridging the language and logical gaps, our system achieves strong results on two datasets featuring realistic utterances (\scholar/ and \geo/).
Despite the fact that our model uses \emph{zero} annotated data for training and validation, it outperforms other supervised methods like \overnight/ and \granno/~\citep{herzig19dontdetect} that require manual annotation.
Analysis shows that current models are far from perfect, suggesting logical gap still remains an issue, while stronger paraphrasers are needed to further close the language gap.

\vspace{-0.2em}
\section{Zero-shot Semantic Parsing via Data Synthesis}
\vspace{-0.2em}
\label{sec:system}
%\label{sec:model:iterative_learning}

%In this section, we introduce our zero-shot semantic parsing model without using annotated training data. 

\paragraph{Problem Definition} Semantic parsers translate a user-issued NL utterance $\utt$ into a machine-executable program $\mr$ (\cref{fig:system}). 
%In this paper we use programs represented as $\lambda$-calculus logical forms~\citep{DBLP:conf/acl/LiangJK11}. \yin{Show an example here}. 
We consider a zero-shot learning setting without access to parallel data in the target domain. 
%, which consists of real-world user-issued natural utterances annotated with programs.
%$\Dnat$
%$\{\langle \uttnat, \mrnat \rangle \}$
Instead, the system is trained on 
%\textbf{can}nonical set $\Dcan$ of 
a collection of machine-synthesized examples.
%, as detailed below.
%$\{\langle \uttcan, \mrcan \rangle \}$.
%The following section presents an overview of the approach.

%\subsection{Learning Semantic Parsers with Synthesized Canonical Data}

\paragraph{Overview} Our system is inspired by the existing zero-shot parser by \citet{Xu2020AutoQA}. 
\cref{fig:system} illustrates our framework.
%illustrative overview of the framework.
Intuitively, we automatically create training examples with canonical utterances from a grammar, which are then paraphrased to increase diversity in language style.
%and the model is then trained on a diverse set of examples with iteratively paraphrased utterances.
%the model generates data by synthesizing programs paired with canonical utterances from a grammar, and then applies iterative rounds of paraphrasing to rewrite the canonical utterances.
Specifically, there are two stages.
First, a set of seed canonical examples (\cref{fig:system:candata}) are generated from a \textbf{synchronous grammar}, which defines compositional rules of NL expressions to form utterances~(\cref{fig:system:grammar}).
Next, in the iterative training stage, a \textbf{paraphrase generation} model rewrites the canonical utterances to more natural and linguistically diverse alternatives~(\cref{fig:system:iter}).
The paraphrased examples are then used to train a semantic parser.
To mitigate noisy paraphrases, a filtering model, which is the parser trained on previous iterations, rejects paraphrases that are potentially incorrect.
This step of paraphrasing and training could proceed for multiple iterations,
with the parser trained on a dataset with growing diversity of language styles.
%This step of paraphrasing and training could repeat iteratively, 
%New canonical examples are then formed by pairing the paraphrased utterances with the original programs.
%To mitigate noise in paraphrasing, a \textbf{filtering} step uses the semantic parser trained on data generated in the previous iteration to parse each paraphrased utterance, and utterances that could be successfully answered by the parser are used to create new samples.
%The paraphrasing (and filtering) step could repeat for multiple iterations, 
%with previously generated examples being further paraphrased, 
%where in each increasing diversity of language styles of the canonical data.
%The paraphrased data is used to train a semantic parser as the final model.

% \begin{figure}[tb]
%   \centering
%   \includegraphics[width=\columnwidth]{figures/grammar.pdf}
%   \caption{Simplified illustration of the grammar rules to generate canonical utterances. {\tt EntSet} denotes entity sets. \yin{TODO: remove ``synchronous'' since we do not show programs here. TODO: This figure is hard to understand. I need to better connect it to the relevant discussions in the text}}
%   \label{fig:grammar}
%\end{figure}

\paragraph{Synchronous Grammar} 
Seed canonical examples are generated from a synchronous context free grammar (SCFG).
%An SCFG is a collection of production rules that define the mapping between logical predicates and their canonical NL expressions.
% (\eg~the abstract entity type {\tt paper}, the database relation {\tt paper.author}, and the entity {\tt author.allan\_turing})
% (\eg~\textit{paper}, \textit{written by}, and \textit{Allan Turing}, resp.)
%\cref{fig:system}(a) lists example productions.
\cref{fig:system:grammar} lists simplified production rules in the grammar.
Intuitively, productions specify how utterances are composed from lower-level language constructs and domain lexicons.
For instance, given a database entity {\tt allan\_turing} with a property {\tt citations}, $\utt_3$ in \cref{fig:system} could be generated using $r_1$.
%A canonical example of synthesized program and utterance (\eg~\textit{paper written by Allan Turing}) could then be formed compositionally using productions in the grammar. 
Productions could be applied recursively to derive more compositional utterances (\eg~$\utt_2$ using $r_2$, $r_4$ and $r_6$).
Our SCFG is based on \citet{herzig19dontdetect}, consisting of domain-general rules of generic logical operations (\eg~{\tt superlative}, $r_3$) and domain-specific lexicons of entity types and relations.
%Specifically, domain-general productions define (1) generic logical operations like {\tt count} and {\tt superlative} (\eg~$r_4$), and (2) 
%compositional rules of English syntax (\eg~$r_1$). 
%compositional rules to construct utterances following English syntax 
%(\eg~a passive verb \textit{written by} is followed by a noun phrase \textit{Dan Klein}).
%Domain-specific rules, on the other hand, are typically used to define task-dependent lexicons like types (\eg~{\tt author}), entities (\eg~{\tt author.allen\_turing}),  and relations (\eg~{\tt author.citations}) in the database. 
%This paper also introduces another type of domain-specific productions, \textbf{idiomatic productions}, which extends the SCFG to cover idiomatic language patterns (\eg~\textit{\underline{Most recent} paper}) tailored to the domain, which we elaborate later in \cref{sec:model:bridge_gap}.
% we generate canonical examples using
Different from \citet{Xu2020AutoQA} which uses a complex grammar with 800 rules, we use a compact grammar with only 46 generic rules plus a handful of  idiomatic productions (\cref{sec:bridge_gaps:idiomatic_rules}) to capture domain-specific language patterns (\eg~\textit{``most recent''} in $\utt_2$, \textit{c.f.},~$\utt_1$).
%This paper also introduces another type of domain-specific productions, \textbf{idiomatic productions} (\cref{sec:bridge_gaps:idiomatic_rules}), which extends the SCFG to capture idiomatic language patterns tailored to the domain (\eg~\textit{``most recent''} in $\utt_2$, \textit{c.f.}~$\utt_1$). 
Given the grammar, examples are enumerated exhaustively up to a threshold of number of rule applications, yielding a large set of seed canonical examples $\Dcan$ (\cref{fig:system:candata}) for paraphrasing.\footnote{SCFGs could not generate utterances with context-dependent rhetorical patterns such as anaphora. Our model could still handle simple domain-specific context-dependent patterns (\eg~\textit{Paper by A and B}, where \textit{A} and \textit{B} are different authors) by first generating all the canonical samples and then filtering those that violate the constraints.}
%, which we elaborate later in .
%\yin{Add one sentence about how we generate the examples from the grammar}

\paragraph{Paraphrase Generation and Filtering} 
The paraphrase generation model rewrites a canonical utterance $\utt$ to more natural and diverse alternatives $\utt'$.
$\utt'$ is then paired with $\utt$'s program to create a new example.
We finetune a \textsc{Bart} model on the dataset by \citet{krishna-etal-2020-reformulating}, a subset of the \textsc{ParaNMT} corpus~\citep{Wieting2018ParaNMT50MPT} that contain lexically and syntactically diverse paraphrases.
The model therefore learns to produce paraphrases with a variety of linguistic patterns, which is essential for closing the language gap when paraphrasing from canonical utterances (\cref{sec:exp}). 
%which encourages the model to produce lexically and syntactically diverse paraphrases.
%As we later show in \cref{sec:exp}, having diverse paraphrases is essential for performance.
%the resulting model learns to generate paraphrases with diverse lexical choices and syntactic structures, which is cruential 
%Our paraphraser is a \textsc{Bart} model specifically fine-tuned to produce lexically and syntactically diverse outputs (\cref{app:paraphraser}).
%Refer to \cref{app:paraphraser} for more details.
Still, some paraphrases are noisy or potentially vague (\filterfn~in \cref{fig:system:iter}). We follow \citet{Xu2020AutoQA} and use the parser trained on previous iterations as the filtering model, and reject paraphrases for which the parser cannot predict their programs.

\vspace*{-0.3em}
\section{Bridging the Gaps between Canonical and Natural Data}
\vspace*{-0.3em}
\label{sec:gaps}
\paragraph{Language and Logical Gaps} The synthesis approach in \cref{sec:system} will generate a large set of paraphrased canonical data (denoted as $\Dpara$). However, as noted by \citet{herzig19dontdetect} (hereafter HB19), the synthetic examples cannot capture all the language and programmatic patterns of real-world natural examples from users (denoted as $\Dnat$).
There are two mismatches between $\Dpara$ and $\Dnat$.
First, there is a \textbf{logical gap} between the programs in $\Dnat$ capturing real user intents, and the synthetic ones in $\Dpara$.
Notably, since programs are exhaustively enumerated from the grammar up to a certain compositional depth, $\Dpara$ will not cover more complex programs in $\Dnat$ beyond the threshold.
Ideally we could improve the coverage using a higher threshold. However, the space of possible programs will grow exponentially, and combinatorial explosion happens even with small thresholds.
% soon becomes intractable. 

Next, there is a \textbf{language gap} between paraphrased canonical utterances and real-world user-issued ones. Real utterances (\eg~the $\utt_2$ in \cref{fig:system},  modeled later in \cref{sec:bridge_gaps:idiomatic_rules}) enjoy more lexical and syntactical diversity, while the auto-paraphrased ones (\eg~$\utt_1'$) are typically biased towards the monotonous and verbose language style of their canonical source (\eg~$\utt_1$).
%For example, the $\utt_1'$ in \cref{fig:system} is still similar to $\utt_1$, whereas a more natural alternative would be $\utt_2$.
While we could increase diversity via iterative rounds of paraphrasing (\eg~$\utt_2 \mapsto \utt_2' \mapsto \utt_2''$), the paraphraser could still fail on canonical utterances that are not natural English sentences at all, like $\utt_1$.

\vspace*{-0.1em}
\subsection{Bridging Language and Logical Gaps}
\vspace*{-0.1em}
\label{sec:bridge_gaps}
We introduce improvements to the system to close the language (\cref{sec:bridge_gaps:idiomatic_rules}) and logical (\cref{sec:bridge_gaps:data_selection}) gaps.

%TBD: Here we describe how we tackle these gaps!
%In this section we elaborate on the improvements on the system described in \cref{sec:system} to close the language (\cref{sec:bridge_gaps:idiomatic_rules}) and logical (\cref{sec:bridge_gaps:data_selection}) gaps.

\vspace*{-0.3em}
\subsubsection{Idiomatic Productions}
\vspace*{-0.3em}
\label{sec:bridge_gaps:idiomatic_rules}

To close language gaps, we augment the grammar with productions capturing   domain-specific idiomatic language styles.
Such productions compress the clunky canonical expressions (\eg~$\utt_1$ in \cref{fig:system}) to more succinct and natural alternatives (\eg~$\utt_2$). We focus on two language patterns:

\paragraph{Non-compositional expressions for multi-hop relations}
Compositional canonical utterances typically feature chained multi-hop relations that are joined together (\eg~\textit{Author that \underline{writes} paper whose \underline{topic} is NLP}), which can be compressed using more succinct phrases to denote the relation chain, where the intermediary pivoting entities (\eg~{\tt paper}) are omitted (\eg~\textit{Author that \underline{works on} NLP}).
The pattern is referred to as sub-lexical compositionality in~\citet{wang15overnight} and used by annotators to compress verbose canonical utterances, while we model them using grammar rules. Refer to \cref{app:grammar} for more details.
%We augment the database with entries for those multi-hop relations (\eg~$\langle${\tt X}, {\tt publish\_in}, {\tt semantic\_parsing}$\rangle$), and then create productions in the grammar aligning those relations with their NL phrases (\eg~\textit{work on}). 

\paragraph{Idiomatic Comparatives and Superlatives}
%Another common scenario where canonical expressions differ much with the idiomatic ones is comparative and superlative constructs.
The general grammar in \cref{fig:system:grammar} uses canonical constructs for comparative (\eg~\textit{smaller than}) and superlative (\eg~\textit{largest}) utterances (\eg~$\utt_1$), which is not ideal for entity types with special units (\eg~time, length).
We therefore create productions specifying idiomatic comparative and superlative expressions (\eg~\textit{paper \underline{published before} 2014}, and $\utt_2$ in \cref{fig:system}).
Sometimes, answering a superlative utterance also requires reasoning with other pivoting entities. For instance, the relation in \textit{``venue that X \underline{publish mostly in}''} between authors and venues implicitly involves counting the papers that \textit{X} publishes.
%More complex programs are required to answer such relations.
%compression of canonical utterances could happen for more complex logical patterns beyond linearly chained relations, and pre-caching the results as assertions in the database could be expensive.
%For example, the relation in \textit{``venue that X \underline{publish mostly in}''} encapsulates a superlative operation and the authorship relation in the canonical utterance \textit{``venue that has \underline{the most number of paper by} X''}).
%TODO: whether to discuss this in the actual paper?
% using productions that map the relation phrases to the logical
For such cases, we create ``macro'' productions, with the NL phrase mapped to a  program that captures the computation involving the pivoting entity (\cref{app:grammar}).
%, instead of a simple DB relation.
%See Appendix xxx for details.

%In line with~\citet{su17crossdomain,Marzoev2020UnnaturalLP}, we remark that such \emph{functionality}-driven grammar engineering to cover patterns in real data is more efficient and cost-effective than example-driven annotation, as SCFG rules are easily comprehensible with basic knowledge of English syntax, and synthetic samples can be further paraphrased to significantly increase linguistic diversity.

\paragraph{Discussion} In line with \citet{su17crossdomain} and \citet{Marzoev2020UnnaturalLP}, we remark that such \emph{functionality}-driven grammar engineering to cover representative patterns in real data using a small set of curated production rules is more efficient and cost-effective than example-driven annotation, 
which requires labeling a sufficient number of parallel samples to effectively train a data-hungry neural model over a variety of underlying meanings and surface language styles.
%which requires manually labeling a large number of parallel samples to cover diverse logical semantics and language styles that is necessary to train a data-hungry neural model.
In contrast, our approach follows \citet{Xu2020AutoQA} to automatically synthesize complex compositional samples from the user-specified productions, which are further paraphrased to significantly increase their linguistic diversity.

\subsubsection{Naturalness-driven Data Selection}
%\yin{@Graham, this might be further reduced, but I've run out of ideas...} \gn{I did some compression, maybe 3-4 lines worth?}\yin{Amazing! I starred at this part for xxxmin and changed nothing..}

\label{sec:bridge_gaps:data_selection}
To cover real programs in $\Dnat$ with complex structures while tackling the exponential sample space, we propose an efficient approach to sub-sample a small set of examples from this space as seed canonical data $\Dcan$ (\cref{fig:system:candata}) for paraphrasing.
Our core idea is to only retain a set of examples $\langle \utt, \mr \rangle$ that most likely reflect the intents of real users.
We use the probability $p_\textrm{LM}(\utt)$ measured by a language model to approximate the ``naturalness'' of canonical examples.\footnote{We use the GPT-2 XL model~\citep{Radford2019LanguageMA}.}
Specifically, given all canonical examples allowed by the grammar, we form buckets based on their derivation depth $d$.
For each bucket $\Dcan^{(d)}$, we compute $p_\textrm{LM}(\utt)$ for its examples, and group the examples using program templates as the key (\eg~$\utt_1$ and $\utt_2$ in \cref{fig:system} are grouped together).
For each group, we find the example $\langle \utt^*, \mr \rangle$ with the highest $p_\textrm{LM}(\utt^*)$, and discard other examples $\langle \utt, \mr \rangle$ if $\log p_\textrm{LM}(\utt^*)\hspace{-0.2mm} -\hspace{-0.2mm} \log p_\textrm{LM}(\utt)\hspace{-1mm} >\hspace{-1mm} \delta$ ($\delta = 5.0$),
removing unlikely utterances from the group (\eg~$\utt_1$).\footnote{$\delta$ chosen in pilot studies, similar to \citet{paws2019naacl}.} 
%, and the remaining utterances in the group could be similarly likely (\eg~``\utterance{Paper by X}'' and ``\utterance{Paper written by X}'').
Finally, we rank all groups in $\Dcan^{(d)}$ based on $p_\textrm{LM}(\utt^*)$, and retain examples in the top-$K$ groups.
%The pruned canonical set therefore contains top-ranked examples for each derivation depth, which are used in subsequent paraphrasing steps.
This method offers trade-off between program coverage and efficiency and, more surprisingly, we show that using only $0.2\%\hspace{-1mm}\sim\hspace{-1mm}1\%$ top-ranked examples also results in significantly better final accuracy (\cref{sec:exp}).
%Our system achieves strong results using only $0.2\%\sim2\%$ top-ranked examples, while including more samples could actually be counter-productive, since most synthesized examples are unlikely (\eg~\textit{Paper by authors that cites authors in deep learning}) and will bring noise to paraphrasing and model optimization.
%$\Dcan$

\vspace*{-0.3em}
\subsection{Generating Validation Data}
\vspace*{-0.3em}
\label{sec:gaps:validation_data}
Zero-shot learning is non-trivial without a high-quality validation set, as the model might overfit on the (paraphrased) canonical data, which is subject to language and logical mismatch.
While existing methods \citep{Xu2020AutoQA} circumvent the issue using real validation data, in this work we create validation sets from paraphrased examples, making our method truly labeled data-free.
Specifically, we consider a two-stage procedure. First, we run the iterative paraphrasing algorithm (\cref{sec:system}) without validation, and then sample $\langle \utt, \mr \rangle$ from its output  with a probability $p(\utt, \mr) \propto p_\textrm{LM}(\utt) ^ \alpha$ ($\alpha=0.4$), ensuring the resulting sampled set $\Dpara^\textrm{val}$ is representative.
%To ensure $\Dpara^\textrm{val}$ contains representative examples, we sample $\langle \utt, \mr \rangle$ from $\Dpara^{0}$ with a probability $p(\utt, \mr) \propto p_\textrm{LM}(\utt) ^ \alpha$ ($\alpha=0.4$).
Second, we restart training using $\Dpara^\textrm{val}$ for validation to find the best checkpoint.
%, where $\alpha$ is a scaling parameter.
The paraphrase filtering model is also initialized with the parser trained in the first stage, which has higher precision and accepts more valid paraphrases.
This is similar to iterative training of weakly-supervised semantic parsers~\citep{Dasigi2019IterativeSF}, where the model searches for candidate programs for unlabeled utterances in multiple stages of learning.

\vspace{-0.5em}
\section{Experiments}
\vspace{-0.5em}
\label{sec:exp}

%In our experiments, we are interested in how the parser performs without manual labels.
%, and in understanding how our methods bridge the language and logical gaps. 
We evaluate our zero-shot parser on two datasets.

\noindent \textbf{\scholar/}~\citep{iyer17user} is a collection of utterances querying an academic database (\cref{fig:system}).
%, with relations defined among entities of authors, papers, and publication venues (\cref{fig:system}).
Examples are collected from users interacting with a parser, which are later augmented with Turker paraphrases.
%paraphrased utterances by crowd workers.
We use the version from \hbpaper~with programs represented in $\lambda$-calculus logical forms. 
The sizes of the train/test splits are 579/211.
Entities in utterances and programs (\eg~{\it \underline{semantic parsing} paper in \underline{ACL}}) are canonicalized to typed slots (\eg~{\tt keyphrase0}, {\tt venue0}) as in \citet{DBLP:conf/acl/DongL16}, and are recovered when programs are executed during evaluation.
We found in the original dataset by \hbpaper, slots are paired with with random entities for execution (\eg~{\tt keyphrase0}$\mapsto$\textit{optics}). Therefore reference programs are likely to execute to empty results, making metrics like answer accuracy more prone to false-positives.
We manually fix all such examples in the dataset, as well as those with execution errors. 
%See Appendix for more details.

\noindent \textbf{\geo/}~\citep{geoquery} is a classical dataset with queries about U.S.~geography (\eg~\utterance{Which rivers run through states bordering California?}). Its database contains basic geographical entities like cities, states, and rivers.
We also use the release from \hbpaper, of size 537/237.

% \vspace*{-0.3em}
% \subsection{Setup}
% \vspace*{-0.3em}
\label{sec:exp:setup}

\paragraph{Models and Configuration} Our semantic parser is a sequence-to-sequence model with a pre-trained \textsc{Bert}$_\textrm{Base}$ encoder~\citep{Devlin2019BERTPO} and an LSTM decoder augmented with a copy mechanism. The paraphraser is a \textsc{Bart}$_\textrm{Large}$ model~\cite{lewis20bart}. 
We use the same set of hyper-parameters for both datasets.
Specifically, we synthesize canonical examples from the SCFG with a maximal program depth of $6$, and collect the top-$K$ ($K=2,000$) GPT-scored sample groups for each depth as the seed canonical data $\Dcan$ (\cref{sec:bridge_gaps:data_selection}). We perform the iterative paraphrasing and training procedure (\cref{sec:system}) for two iterations.
%, where the parser is trained for 30 epochs in each iteration. 
%In each iteration, we train the parser for 30 epochs. 
%The beam size for the paraphraser is 20. 
We create validation sets of size $2,000$ in the first stage of learning (\cref{sec:gaps:validation_data}), and perform validation using perplexity in the second stage. 
Refer to \cref{app:config} for more details.
Note that our model only uses the natural examples in both datasets for evaluation purposes, and the training and validation splits are \emph{not} used during learning.

\paragraph{Measuring Language and Logical Gaps}
We measure the language mismatch between utterances in the paraphrased canonical ($\Dpara$) and natural ($\Dnat$) data using \textbf{perplexities} of natural utterances in $\Dnat$ given by a GPT-2 LM fine-tuned on $\Dpara$.
For logical gap, we follow \hbpaper~and compute the \textbf{coverage} of natural programs $\mr \in \Dnat$ in $\Dpara$. 

\paragraph{Metric} We use \textbf{denotation accuracy} on the execution results of model-predicted programs. We report the mean and standard deviation with five random restarts.
%We ran all experiments with five random restarts.

\vspace*{-0.3em}
\subsection{Results}
\vspace*{-0.3em}
\label{sec:exp:result}

%\jw{More sign posting here might help - there is a lot of information on different topics so starting this with an overview of what you'll be presenting here may make it easier for readers to keep track of all your findings.}
%In experiments, 
We first compare our model with existing approaches using labeled data. Next, we analyze how our proposed methods close the language and logical gaps.
%First, we compare our zero-shot parser with existing semantic parsing models that require labeled data.
\cref{tab:exp:system_comparison} reports accuracies of various systems on the test sets, as well as their form of supervision.
%We compare with existing systems which require manually labeled training data.
Specifically, the \textbf{supervised} parser uses a standard parallel corpus $\mathbb{D}_\textrm{nat}$ of real utterances annotated with programs.
\textbf{\overnight/} uses paraphrased synthetic examples $\Dpara$ like our model, but with manually written paraphrases.
\textbf{\granno/} uses unlabeled real utterances $\uttnat \in \Dnat$, and manual paraphrase detection to pair $\uttnat$ with the canonical examples $\Dcan$.
%the a sub-set of $\Dpara$, consisting of examples where annotators are able to label their utterances \uttcan~as paraphrases of some natural utterances in $\uttnat \in \Dnat$, and replace \uttcan~with \uttnat~for training.
%Our model does not use any annotated data, while 
Our model outperforms existing approaches on the two benchmarks 
%achieves competitive performance on the two benchmarks, matching or outperforming \granno/ on \geo/ ($72.2$ vs.~$72.0$) and \scholar/ ($72.5$ vs.~$69.2$), respectively.
without using any annotated data, while \granno/, the currently most cost-effective approach, still spends \$155 in manual annotation (besides collecting real utterances) to create training data for the two datasets~(\newcite{herzig19dontdetect}, \hbpaper).
Overall, the results demonstrate that zero-shot parser based on idiomatic synchronous grammars and automatic paraphrasing using pre-trained LMs is a data-efficient and cost-effective paradigm to train semantic parsers for emerging domains.

\begin{table}[t]
    \centering
    \small
    \def\arraystretch{1}%
    \setlength{\tabcolsep}{2pt}
    % \begin{tabular}{lp{3.2cm}cc}
    % \toprule
    %     System & Supervision & \scholar/ & \geo/ \\ \midrule
    %     Supervised$^\dagger$  & Labeled Examples & 83.4 & 86.3  \\ \hdashline
    %     \overnight/$^\dagger$ & Manual Paraphrases & 40.8 & 61.9  \\
    %     \granno/$^\dagger$    & Real Utterances, Manual Paraphrase Detection & 69.2 & 72.0 \\ 
    %     %Our System & 72.2 & 72.5 \\ 
    %     %Our System  & \textbf{74.6} & \textbf{75.2} \\
    %     Our System & $-$ & \textbf{75.5} \std{1.6} & \textbf{74.1} \std{2.3} \\ 
    %     \bottomrule
    % \end{tabular}
    \resizebox{\columnwidth}{!}{
    \begin{tabular}{lp{3.2cm}cc}
    \toprule
        System & Supervision & \scholar/ & \geo/ \\ \midrule
        Supervised$^\dagger$  & Labeled Examples & 79.7 \std{2.2} & 81.9 \std{5.3}  \\ \hdashline
        \overnight/$^\dagger$ & Manual Paraphrases & 41.0 \std{3.8} & 55.8 \std{6.4}  \\
        \granno/$^\dagger$    & Real Utterances, Manual Paraphrase Detection & 68.3 \std{1.6} & 69.4 \std{1.9} \\ 
        %Our System & 72.2 & 72.5 \\ 
        %Our System  & \textbf{74.6} & \textbf{75.2} \\
        Our System & $-$ & \textbf{75.5} \std{1.6} & \textbf{74.1} \std{2.3} \\ 
        \bottomrule
    \end{tabular}}
    \caption{Averaged denotation accuracy and standard deviation on \textsc{Test} sets. Results are averaged with five random restarts. $^\dagger$Models originally from \citet{herzig19dontdetect} and run with five random restarts. Results from our model are tested \textit{v.s.}~\textsc{Granno} using paired permutation test with $p < 0.05$.}
    % $^\dagger$Results reported in \citet{herzig19dontdetect}.
    % 
    \label{tab:exp:system_comparison}
    \vspace{-5mm}
\end{table}

% Still, our system falls behind fully supervised models trained on natural datasets $\Dnat$, due to language and logical gaps between the auto-paraphrased canonical data $\Dpara$ and $\Dnat$.
Still, our system falls behind fully supervised models trained on natural datasets $\Dnat$, due to language and logical gaps between $\Dpara$ and $\Dnat$.
%Later parts of this section will provide more insights into such language and logical gaps faced by zero-shot semantic parsers.
%In following experiments, we attempt to understand the impact of such gaps on the system's performance. 
%Specifically, we explore whether our proposed methods are effective at narrowing the gaps and improving accuracy.
%Since the commonly-used validation splits of the two benchmarks are quite small (\eg~only 99 samples for \scholar/), in the following we use the full training (and validation) splits for evaluation to get more reliable results.
In following experiments, we explore whether our proposed methods are effective at narrowing the gaps and improving accuracy. 
Since the validation splits of the two datasets are small (\eg~only 99 samples for \scholar/), we use the full training/validation splits for evaluation to get more reliable results.

\paragraph{More expressive grammars narrow language and logical gaps}
We capture domain-specific language patterns using idiomatic productions to close language mismatch (\cref{sec:bridge_gaps:idiomatic_rules}). 
\cref{tab:exp:scholar:lang_ablation,tab:exp:geo:lang_ablation} list the results when we gradually improve the expressiveness of the grammar by adding different types of  idiomatic productions.
We observe more expressive grammars help close the language gap, as indicated by the decreasing perplexities. 
%on real utterances.
%This is especially important for \scholar/, which features diverse idiomatic language patterns that are hard to infer from plain canonical utterances without using idiomatic productions.
This is especially important for \scholar/, which features diverse idiomatic NL expressions hard to infer from plain canonical utterances.
%For instance, it could be non-trivial to paraphrase superlative canonical utterances like \utterance{Topic of the most number of ACL paper} to more idiomatic ones (\eg~\utterance{\underline{The most popular topic for} ACL paper}), while directly including such superlative relations in the grammar ($+$Superlative) could be helpful.
For instance, it could be non-trivial to paraphrase canonical utterances with multi-hop (\eg~\utterance{Author that \underline{cites} paper \underline{by} X}) or superlative relations (\eg~\utterance{Topic of \underline{the most number of} ACL paper}) to more idiomatic alternatives (\eg~``\utterance{Author that \underline{cites} X}'', and ``\utterance{\underline{The most popular topic for} ACL paper}''), while directly including such patterns in the grammar (\textbf{$+$Multihop Rel.}~and \textbf{$+$Superlative}) is helpful. 

\begin{table}[t]
    \centering
    \small
    \begin{tabular}{lcccc}
    \toprule
    \multirow{2}{*}{Grammar}   & \multirow{2}{*}{\textsc{Acc.}} & \multirow{2}{*}{PPL} & \multicolumn{2}{c}{Logical Coverage} \\ 
    & & & $\Dcan$ & $\Dpara$ \\ \midrule
    Base              & 66.3 \std{3.7}    & 23.0   & 80.6 & 75.8 \\
    $+$Multihop Rel.  & 67.0 \std{1.1}    & 22.0   & 87.7 & 81.2 \\
    $+$Comparison     & 67.3 \std{2.4}    & 21.7   & 86.5 & 80.2 \\
    $+$Superlative    & 77.8 \std{2.2}    & 20.9   & 90.6 & 86.1 \\ 
      ~~~$-$Multihop Rel. & 75.8 \std{3.4} & 20.8  & 83.9 & 81.1 \\ \bottomrule
    %$+$Compound Noun  & 76.9    & 18.7/16.9   & $I:$TBD $E:$88.0 \\ \bottomrule
    \end{tabular}
    \caption{Ablation of grammar categories on \scholar/.}
    \label{tab:exp:scholar:lang_ablation}
    \vspace{-3mm}
\end{table}

\begin{table}[t]
    \centering
    \small
    \begin{tabular}{lcccc}
        \toprule
        \multirow{2}{*}{Grammar}   & \multirow{2}{*}{\textsc{Acc.}} & \multirow{2}{*}{PPL} & \multicolumn{2}{c}{Logical Coverage} \\ 
        & & & $\Dcan$ & $\Dpara$ \\ \midrule
        %Base              & TBD & TBD & TBD \\
        Base            & 64.5 \std{4.6}  & 8.2   & 84.4 & 79.7      \\
        $+$Multihop Rel.& 67.9 \std{4.0}  & 8.1   & 83.6 & 79.7      \\
        $+$Superlative  & 72.8 \std{2.8} & 8.0   & 84.1 & 79.4   \\
        ~~~$-$Multihop Rel. & 66.5 \std{3.7} & 8.2  & 84.1 & 80.0 \\ \bottomrule
    \end{tabular}
    \caption{Ablation study of grammar categories on \geo/.}
    \label{tab:exp:geo:lang_ablation}
    \vspace{-5mm}
\end{table}

Additionally, we observe that more expressive grammars also improve logical coverage.
The last columns (\textbf{Logical Coverage}) of \cref{tab:exp:scholar:lang_ablation,tab:exp:geo:lang_ablation} report the percentage of real programs that are covered by the seed canonical data before ($\Dcan$) and after ($\Dpara$) iterative paraphrasing.
%both for the initially synthesized examples ($I$, (a) in \cref{fig:system}) and the final data after iterative paraphrasing ($E$, (b) in \cref{fig:system}).
Intuitively, idiomatic grammar rules could capture compositional program patterns like multi-hop relations 
%(\eg~\utterance{Author that \underline{cites} paper \underline{written by} X}) 
and complex superlative queries (\eg~\utterance{Author that \underline{publish mostly in} ACL}, \cref{sec:bridge_gaps:idiomatic_rules}) within a single production, enabling the grammar to generate more compositional programs under the same threshold on the derivation depth. 
% the number of rule applications. 
Notably, when adding all the idiomatic productions on \scholar/, the number of exhaustively generated examples with a program depth of $6$ is tripled ($530K\mapsto1,700K$).
%more compositional programs could then be generated from the grammar given the same threshold on the number of rule applications. 
%(\eg~a {\tt CountSuperlative} function on {\tt citation\_count)}), 
% which could fall out of the threshold. 

Moreover, recall that the seed canonical dataset $\Dcan$ contains examples with highly-likely utterances under the LM (\cref{sec:bridge_gaps:data_selection}).
%using examples with highest-scored utterances ranked by LMs (\cref{sec:bridge_gaps:data_selection}).
%selects the top-ranked canonical examples scored by GPT-2 as training instances. 
Therefore, examples created by idiomatic productions are more likely to be included in $\Dcan$, as their more natural and well-formed utterances often have higher LM scores.
%as their utterances are more natural and could score higher.
%This also helps improve logical coverage, as idiomatic productions are intended to capture programmatic patterns in real data.
%However, we also remark that while examples generated with idiomatic productions are more likely to reflect logical patterns in real data, such examples could over dominate the shorted-listed canonical sample set after LM filtering, leaving other useful examples with lower LM scores out.
However, note that this could also be counter-productive, as examples created with idiomatic productions could dominate the LM-filtered $\Dcan$, ``crowding out'' other useful examples with lower LM scores.
This likely explains the slightly decreased logical coverage on \geo/ (\cref{tab:exp:geo:lang_ablation}), as more than 30\% samples in the filtered $\Dcan$ include idiomatic multi-hop relations directly connecting geographic entities with their countries (\eg~``\utterance{City \underline{in} US}'', \textit{c.f.}~``\utterance{City \underline{in} state \underline{in} US}''), while such examples only account for $\sim8\%$ of real data.
%While this over-representation issue of some idiomatic relations might not negatively impact accuracy, we leave exploring generation of more balanced canonical data as important future work.
While the over-representation issue might not negatively impact accuracy, we leave generating more balanced synthetic data as important future work.
%  and superlative relations (\eg~\utterance{The \underline{longest} river in California})
%With these factors combined, after adding superlative relations ($+$Superlative), empirically we observe the grammar yields \yin{xxx\%} more examples with program depth larger than 5 on \scholar/.

%For instance, users are likely to ask utterances like $\utt:$\textit{Most recent paper by Alan Turing}, but its the canonical example \textit{Venue of the paper by Alan Turing and that has the largest publication year} fails to be included in the training set because of low LM probability.
%With domain-specific grammar rules for superlative queries, examples with the more idiomatic utterance $\utt$ are covered by the training data.
Finally, we note that the logical coverage drops after paraphrasing ($\Dcan$ \textit{v.s.}~$\Dpara$ in \cref{tab:exp:scholar:lang_ablation,tab:exp:geo:lang_ablation}). This is because for some samples in $\Dcan$, the paraphrase filtering model rejects all their paraphrases. 
We provide further analysis later in a case study.
%This calls for better paraphrase filtering models, which we provide further analysis later in this section.

\begin{table*}[t]
    \small
    \centering
    \begin{tabular}{llcccccc:cc:cc}
    \toprule
   & \multirow{2}{*}{$K$} & \multicolumn{2}{c}{Train Data Size} & \multirow{2}{*}{\textsc{Acc}} & \multirow{2}{*}{PPL} & \multicolumn{2}{c}{Logical Coverage} & \multicolumn{2}{c}{In Coverage} & \multicolumn{2}{c}{Out of Coverage} \\ 
   & & $|\Dcan|$ & $|\Dpara|$ & & & $\Dcan$ & $\Dpara$ & \textsc{Acc} & PPL & \textsc{Acc} & PPL \\ \midrule
\multirow{5}{*}{\rotatebox{90}{\scholar/}} 
   & 500  & 1,554 & 45,269   & 74.0 \std{3.7} & 22.0 & 79.4 & 76.0~\notenum{14.5} & 82.3 & 23.4 & 47.6 & 18.2 \\
   & 1,000 & 3,129 & 80,481   & 75.9 \std{1.7} & 21.4 & 88.0 & 82.0~\notenum{9.4}  & 81.4 & 21.3 & 50.6 & 21.7 \\
   & 2,000 & 5,486 & 129,955  & 77.8 \std{2.2} & 20.9 & 90.6 & 86.1~\notenum{7.5}  & 82.2 & 20.7 & 50.2 & 22.7 \\
   & 4,000 & 9,239 & 202,429  & 78.4 \std{1.7} & 20.7 & 91.9 & 87.4~\notenum{4.9}  & 83.2 & 20.5 & 45.3 & 22.0 \\
   & 8,000 & 17,077 & 330,548 & 75.5 \std{2.1} & 21.5 & 92.0 & 88.2~\notenum{2.9}  & 79.8 & 21.4 & 43.4 & 22.4 \\ \midrule
\multirow{5}{*}{\rotatebox{90}{\geo/}} 
   & 500  & 1,351 & 29,835   & 61.6 \std{5.4} & 8.4 & 70.3 & 64.4~\notenum{14.2} & 79.2 & 7.6 & 29.8 & 9.9  \\
   & 1,000 & 2,586 & 55,117   & 68.5 \std{7.7} & 8.2 & 80.5 & 74.9~\notenum{9.0}  & 81.4 & 7.4 & 28.8 & 11.3 \\
   & 2,000 & 5,413 & 112,530  & 72.8 \std{2.8} & 8.0 & 84.1 & 79.4~\notenum{5.2}  & 82.0 & 7.4 & 37.6 & 10.8 \\
   & 4,000 & 11,085 & 182,469 & 67.5 \std{6.3} & 8.2 & 84.9 & 78.3~\notenum{3.1}  & 75.5 & 7.6 & 38.8 & 11.2 \\
   & 8,000 & 16,312 & 243,343 & 67.9 \std{4.5} & 8.2 & 85.4 & 78.0~\notenum{2.1}  & 75.5 & 7.5 & 41.3 & 11.2 \\ \bottomrule
    \end{tabular}
    \caption{Results on \scholar/ and \geo/ with varying amount of canonical examples in the seed training data.}
    \label{exp:tab:scholar_geo_with_train_size_main}
    \vspace{-3mm}
\end{table*}

\paragraph{Do smaller logical gaps entail better performance?} 
%We note that these two factors are not independent. 
%Therefore we perform controlled experiments using different configurations of the system, and study the correlation between quantitative changes of the gaps and the accuracies on natural evaluation data.
As in \cref{sec:bridge_gaps:data_selection}, the seed canonical data $\Dfcan$ consists of top-$K$ highest-scoring examples under GPT-2 for each program depth.
This data selection method makes it possible to train the model efficiently in the iterative paraphrasing stage using a small set of canonical samples that most likely appear in natural data out of the exponentially large sample space.
However, using a smaller cutoff threshold $K$ might sacrifice logical coverage, as fewer examples are in $\Dfcan$.
%in $\Dcan$, leading to larger logical gap. 
To investigate this trade-off, we report results with varying $K$ in \cref{exp:tab:scholar_geo_with_train_size_main}.
Notably, with $K=1,000$ and around $3K$ seed canonical data $ \Dfcan $ (before iterative paraphrasing), $ \Dfcan $ already covers $88\%$ and $80\%$ natural programs on \scholar/ and \geo/, resp.
This small portion of samples only account for $0.2\%$ ($1\%$) of the full set of $1.7M+$ ($0.27M$) canonical examples exhaustively generated from the grammar on \scholar/ (\geo/).
This demonstrates our data selection approach is effective in maintaining learning efficiency 
%with smaller size of training data 
while closing the logical gap.
In contrast, the baseline data selection strategy of randomly choosing canonical examples from each level of program depth instead of using the top-$K$ highest scored samples is less effective. As an example, this baseline strategy achieves an accuracy of 69.7\% and 65.5\% on \scholar/ and \geo/ respectively when $K=2,000$, which is around $7\%$ lower than the accuracy achieved by our approach ($77.8\%$ and $72.8\%$, \cref{exp:tab:scholar_geo_with_train_size_main}).
%compared to $77.8\%$ and $72.8\%$ registered by our method.

More interestingly, while larger $K$ yields higher logical form coverage, the accuracy might not improve.
% even with higher logical form coverage w
%When $K > 2000$, the performance on natural data starts to drop even with higher logical form coverage.
This is possibly because while the recall of real programs improves, the percentage of such programs in paraphrased canonical data $\Dpara$ (numbers in parentheses) actually drops.
Out of the remaining $90\%+$ samples in $\Dpara$ whose programs are not in $\Dnat$, many have unnatural intents that real users are unlikely to issue (\eg~``\utterance{Number of titles of papers with the smallest citations}'', or ``\utterance{Mountain whose elevation is the length of Colorado River}''). 
%utterances  set of additional samples whose programs are not included in the natural data are unlikely to appear 
%This result might imply that the additional unlikely canonical training examples whose programs are not in the natural data (\eg~\textit{Mountain whose elevation is lower than the length of Colorado River.}) could be harmful to the model.
Such \emph{unlikely} samples are potentially harmful to the model, causing worse language mismatch, as suggested by the increasing perplexity when $K = 8,000$.
%This is also reflected in the increased perplexity, which suggests the paraphrases of those unlikely samples 
Similar to \hbpaper, empirically we observe around one-third of samples in $\Dcan$ and $\Dpara$ are unlikely.
%\citet{herzig19dontdetect} report around $31\%$ samples generated from a canonical grammar are unlikely.
%In our scenario with more expressive grammars, we observe that $33\%$ and \yin{xx\%} canonical examples (out of 100 random samples) are unlikely on the two benchmarks.
As later in the case study, such unlikely utterances have noisier paraphrases, which hurts the quality of $\Dpara$.
%To better understand the impact of such unlikely examples to the paraphrasing and learning process, we performed a controlled experiments...\yin{TODO:think of what kind of exp to add here...}

\paragraph{Does the model generalize to out-of-distribution samples?}
Next, to investigate whether the model could generalize to utterances with out-of-distribution program patterns not seen in the training data $\Dpara$, we report accuracies on the splits whose program templates are covered (\textbf{In Coverage}) and not covered (\textbf{Out of Coverage}) by $\Dpara$.
%in \cref{tab:exp:scholar:results_with_data_size,tab:exp:geo:result_with_data_size}. 
%, as listed in the ``In Coverage'' (``Out of Coverage'') columns in \cref{tab:exp:scholar:results_with_data_size,tab:exp:geo:result_with_data_size}.
%Not surprisingly, the model performs significantly better on the in-coverage set, with generally lower perplexity, indicating less language mismatch.
Not surprisingly, the model performs significantly better on the in-coverage sets with less language mismatch.\footnote{An exception is $K\hspace{-1mm}=\hspace{-1mm}500$ on \scholar/, where the perplexity on out-of-coverage samples is slightly lower.
This is because utterances in \scholar/ tend to use compound nouns to specify compositional constraints (\eg~\textit{\underline{ACL} \underline{2021} \underline{parsing} papers}), a language style common for in-coverage samples but not captured by the grammar.
With smaller $K$ and $\Dcan$, it is less likely for the paraphrased data $\Dpara$ to capture similar syntactic patterns.
Anther factor that makes the out-of-coverage PPL smaller when $K=500$ is that there are more (simpler) examples in the set compared to $K>500$, and the relatively simple utterances will also bring down the PPL.}
%With small $K$ and fewer examples in $\Dcan$, it is less likely for the subsequent paraphrased data $\Dpara$ to contain utterances with similar syntactic structures.
%, leading to higher perplexity. 
%Still, our model based on a \textsc{Bert} encoder could handle such utterances with novel language patterns.
Our results are also in line with recent research in compositional generalization of semantic parsers~\citep{lake18generalization,finegan18improving}, which suggests that existing models generalize poorly to utterances with novel compositional patterns (\eg~conjunctive objects like \utterance{Most cited paper by \underline{X and Y}}) not seen during training.
% (\eg~training examples only have single names as objects).
%\citet{herzig19dontdetect} also reports similar findings for \overnight/ models using synthesized examples with manual paraphrases. 
Still surprisingly, our model generalizes reasonably to compositionally novel (out-of-coverage) splits, registering $30\%\hspace{-1mm}\sim\hspace{-1mm}50\%$ accuracies, in contrast to \hbpaper~reporting accuracies smaller than $10\%$ on similar benchmarks for \overnight/.
%We hypothesize that auto-paraphrasing of significantly larger amounts of compositional examples 
We hypothesize that synthesizing compositional samples increases the number of unique program templates in training, which could be helpful for compositional generalization~\citep{Akyrek2020LearningTR}. 
As an example, the number of unique program templates in $\Dpara$ when $K=2,000$ on \scholar/ and \geo/ is $1.9K$ and $1.7K$, resp, compared to only $125$ and $187$ in $\Dnat$.
This finding is reminiscent of data augmentation strategies for supervised parsers using synthetic samples induced from (annotated) parallel data~\citep{Jia2016,Wang2021LearningTS}.

% \begin{table}[t]
%     \centering
%     \small
%     \begin{tabular}{lcc}
%     \toprule
%       Paraphraser & \scholar/ & \geo/ \\ \midrule
%       This work & 78.2 & 73.8 \\
%       \citet{Xu2020AutoQA} & 73.5 & 67.0 \\
%     \bottomrule
%     \end{tabular}
%     \caption{Denotation accuracies on \textsc{Dev} sets with different paraphrasers.\yin{Xu2020 model has a weird bug..... The results might not reflect the actual results....}}
%     \label{tab:exp:paraphrasers}
% \end{table}

\paragraph{Impact of Validation Data}
Our system generates validation data from samples of the paraphrased data in an initial run  (\cref{sec:gaps:validation_data}). Tab~\cref{tab:exp:dev_splits} compares this strategy of generating validation data with a baseline approach, which randomly splits the seed canonical examples in $\Dcan$ into training and validation sets, and runs the iterative paraphrasing and training algorithm on the two sets in parallel. In each iteration, the checkpoint that achieves the best perplexity on the paraphrased validation examples is saved. We use the paraphrase filtering model learned on the training set to filter the paraphrases of validation examples. 
This baseline approach performs reasonably well. Still, empirically we find this strategy creates larger logical gaps, as some canonical samples whose program patterns appear in the natural data $\Dnat$ could be partitioned into the validation data, and not used for training.

\begin{table}[t]
  \small
  \centering
  \begin{tabular}{lcc}
    \toprule
    Validation Method             & \scholar/ &  \geo/  \\ \midrule
    Our Approach (\cref{sec:gaps:validation_data})            & 77.8 \std{2.2} & 72.8 \std{2.8} \\
    Random Split of $\Dcan$     & 74.1 \std{1.5} & 69.7 \std{3.3} \\ \bottomrule
    %\textit{template} Split of $\Dcan$  &     \\
  \end{tabular}
    \caption{Denotation accuracies of different strategies to generate validation data.}
    \label{tab:exp:dev_splits}
    \vspace{-3mm}
\end{table}
\begin{table}[t]
\def\arraystretch{1}%
\setlength{\tabcolsep}{2pt}
\resizebox{\columnwidth}{!}{
\begin{tabular}{lccc:ccc}
\toprule
\multirow{2}{*}{Paraphraser} & \multicolumn{3}{c}{\scholar/} & \multicolumn{3}{c}{\geo/}   \\ 
                             & Tok.~$F_1$$\downarrow$  & $\tau$$\downarrow$  & \textsc{Acc.}$\uparrow$ & Tok.~$F_1$$\downarrow$ & $\tau$$\downarrow$ & \textsc{Acc.}$\uparrow$ \\ \midrule
Ours                    & 70.3  & 0.71  & 77.8 & 69.2       & 0.78    & 72.8 \\
\citet{Xu2020AutoQA}    & 72.4  & 0.94  & 69.9 & 74.5       & 0.95    & 62.3  \\ \bottomrule
\end{tabular}}
\caption{Systems with different paraphrasers. We report end-to-end denotation accuracy, as well as $F_1$ and Kendall's $\tau$ rank coefficient between utterances and their paraphrases.}
\label{tab:exp:paraphrasers}
\vspace*{-5mm}
\end{table}

\paragraph{Impact of Paraphrasers} Our system relies on strong paraphrasers to generate diverse utterances in order to close the language gap. \cref{tab:exp:paraphrasers} compares the performance of the system trained with our paraphraser and the one used in \citet{Xu2020AutoQA}.
Both models are based on \textsc{Bart}, while our paraphraser is fine-tuned to encourage lexically and syntactically diverse outputs (\cref{app:paraphraser}).
We measure lexical diversity using token-level $F_1$ between the original and paraphrased utterances $\langle \utt, \utt' \rangle$~\citep{Rajpurkar2016SQuAD10,krishna-etal-2020-reformulating}. 
For syntactic divergence, we use Kendall's $\tau$~\citep{lapata-2006-automatic} to compute the ordinal correlation between $\utt$ and $\utt'$, which intuitively measures the number of times to swap tokens in $\utt$ to get $\utt'$ using bubble sort.
Our paraphraser generates more diverse paraphrases (\eg~\utterance{What is the biggest state in US?}) from the source (\eg~\utterance{State in US and that has the largest area}), as indicated by lower token-level overlaps and ordinal coefficients, comparing to the existing paraphraser (\eg~\utterance{The state in US with the largest surface area}).
Nevertheless, our paraphraser is still not perfect, as discussed next.
%as suggested by lower token-level overlap and order correlation, resulting in better end-to-end performance.

% \begin{itemize}
%     \item Results using Oracle Programs and Languages
%     \item Results using Oracle Paraphrase ID model.
% \end{itemize}

\begin{table}[t]
  \small
  %\centering
  \def\arraystretch{1}%
  \setlength{\tabcolsep}{1pt}
  \resizebox{1.03 \columnwidth}{!}{
  \begin{tabular}{lp{7.8cm}}
  \hline
    \rowcolor{black!10!} \multicolumn{2}{c}{Example 1 (Uncommon Concept)} \\
    $\utt_1$  & \textit{Venue of paper by \textnormal{\tt author}$_0$ and published in \textnormal{\tt year}$_0$} \\ 
    $\utt'_{1,1}$ & \textit{\textnormal{\tt author}$_0$'s paper, published in \textnormal{\tt year}$_0$} ~~ \xmark \\
    $\utt'_{1,2}$ & \textit{Where the paper was published by \textnormal{\tt author}$_0$ in \textnormal{\tt year}$_0$?} \filterfn \\
    $\utt'_{1,3}$ & \textit{Where the paper was published in \textnormal{\tt year}$_0$ by \textnormal{\tt author}$_0$?} \filterfn \\ \hdashline
    $\uttnat^*$  & \textit{Where did \textnormal{\tt author}$_0$ publish in \textnormal{\tt year}$_0$?} \hfill \textcolor{red}{(Wrong Answer)} \\ \hline
    
    \rowcolor{black!10!} \multicolumn{2}{c}{Example 2 (Novel Language Pattern)} \\
    $\utt_2$   & \textit{Author of paper published in \textnormal{\tt venue}$_0$ and in \textnormal{\tt year}$_0$} \\ 
    $\utt'_{2,1}$  & \textit{Author of papers published in \textnormal{\tt venue}$_0$ in \textnormal{\tt year}$_0$} ~~ \cmark \\
    $\utt'_{2,2}$  & \textit{Who wrote a paper for \textnormal{\tt venue}$_0$ in \textnormal{\tt year}$_0$} ~~ \filterfn \\
    $\utt'_{2,3}$  & \textit{Who wrote the \textnormal{\tt venue}$_0$ paper in \textnormal{\tt year}$_0$} ~~ \filterfn \\ \hdashline
    $\uttnat^*$   & \textit{\textnormal{\tt venue}$_0$ \textnormal{\tt year}$_0$ authors} \hfill \textcolor{deepgreen}{(Correct)} \\ \hline

    \rowcolor{black!10!} \multicolumn{2}{c}{Example 3 (Unnatural Utterance)} \\
    $\utt_3$  & \textit{Author of paper by \textnormal{\tt author}$_0$} \\
    $\utt'_{3,1}$ & \textit{Author of the paper written by \textnormal{\tt author}$_0$} ~~ \cmark \\
    $\utt'_{3,2}$ & \textit{Author of \textnormal{\tt author}$_0$'s paper} ~~ \cmark \\
    $\utt'_{3,3}$ & \textit{Who wrote the paper \textnormal{\tt author}$_0$ wrote?} ~~ \filterfn \\ \hdashline
    $\uttnat^*$   & \textit{Co-authors of \textnormal{\tt author}$_0$} \hfill \textcolor{red}{(Wrong Answer)} \\ \hline

    \rowcolor{black!10!} \multicolumn{2}{c}{Example 4 (Unlikely Example)} \\
    $\utt_4$   & \textit{Paper in \textnormal{\tt year}$_0$ and whose author is not the most cited author} \\ 
    $\utt'_{4,1}$  & \textit{A paper published in \textnormal{\tt year}$_0$ that isn't the most cited author} \filterfp \\
    $\utt'_{4,2}$  & \textit{What's not the most cited author in \textnormal{\tt year}$_0$} \filterfp \\
    $\utt'_{4,3}$  & \textit{In \textnormal{\tt year}$_0$, he was not the most cited author} \filterfp \\ 

  \bottomrule
  \end{tabular}
  }
  \caption{Case Study on \scholar/. We show the seed canonical utterance $\utt_i$, the paraphrases $\utt'_{i, j}$, and the relevant natural examples $\utt^*_\textrm{nat}$. \cmark~and \xmark~denote the correctness of paraphrases. \filterfn~denotes false negatives of the filtering model (correct paraphrases that are filtered), \filterfp~denotes false positives (incorrect paraphrases that are accepted). Entities are canonicalized with indexed}
  \label{tab:exp:case_study}
  \vspace*{-6mm}
\end{table}

\vspace{-0.5em}
\section{Limitations and Discussion}
\vspace{-0.5em}
Our parser still lags behind the fully supervised model (\cref{tab:exp:system_comparison}).
To understand the remaining bottlenecks, we show representative examples in \cref{tab:exp:case_study}.

\paragraph{Low Recall of Filter Model}
%While our model outperforms other annotation-efficient approaches, it 
%Our parser still lags behind the fully supervised model (\cref{tab:exp:system_comparison}).
%To understand the remaining bottlenecks, we show representative examples in \cref{tab:exp:case_study}.
%We are particularly interested in cases where the program pattern of a natural utterance is covered by the seed canonical data $\Dcan$, as fixing those examples is could be relatively easier compared to improving logical coverage.
First, the recall of the paraphrase filtering model is low.
%For these cases, the major cause of error is the poor recall of the paraphrase filtering model, which leverages the semantic parser trained on the canonical data generated in previous iterations.
The filtering model uses the parser trained on the paraphrased data generated in previous iterations.
Since this model is less accurate, it can incorrectly reject valid paraphrases $\utt'$ (\filterfn~in \cref{tab:exp:case_study}), especially when $\utt'$ uses a different sentence type (\eg~questions) than the source (\eg~statements).
Empirically, we found the recall of the filtering model at the first iteration of the second-stage training (\cref{sec:gaps:validation_data}) is only around $60\%$.
This creates logical gaps, as paraphrases of examples in the seed canonical data $\Dcan$ could be rejected by the conservative filtering model, leaving no samples with the same programs in $\Dpara$.
%, suggesting that better paraphrase identification model is the key to improve the current approach.
%To understand the potential improvements when switching to a better filtering model, we designed an ``oracle'' filterer, using a RoBERTa model fine-tuned on a parallel corpus of natural utterances $\uttnat$ and auto-paraphrased one $\uttcan$ where $\uttnat$ 

\paragraph{Imperfect Paraphraser}
The imperfect paraphraser could generate semantically incorrect predictions (\eg~$\utt'_{1,1}$), especially when the source canonical utterance contains uncommon or polysemic concepts (\eg~\textit{venue} in $\utt_1$), which tend to be ignored or interpreted as other entities (\eg~\textit{sites}).
%Besides the lack of coverage for idiomatic relations, utterances can also follow special compositionality patterns. 
Besides rare concepts, the paraphraser could also fail on utterances that follow special compositionality patterns.
For instance, $\uttnat^*$ in Example 2 uses compound nouns to denote the occurrence of a conference, which is difficult to automatically paraphrase from $\utt_2$ (that uses prepositional phrases) without any domain knowledge. While the model could still correctly answer $\uttnat^*$ in this case, $\uttnat^*$'s perplexity is high, suggesting language mismatch.

\paragraph{Unnatural Utterances}
%\paragraph{Limited Coverage of SCFG}
While we have attempted to close the language gap by generating canonical utterances that are more idiomatic in language style, some of those synthetic utterances are still not natural enough for the paraphraser to rewrite.
This is especially problematic for relations not covered by our idiomatic productions.
For instance, our SCFG does not cover the co-authorship relation in Example 3. Therefore the generated synthetic utterance $\utt_3$ uses a clumsy multi-hop query to express this intent, which is non-trivial for the model to paraphrase to an idiomatic expression such as $\uttnat^*$.
While this issue could be potentially mitigated using additional production rules, grammar engineering could still remain challenging, as elaborated later in this section. 

\paragraph{Unlikely Examples}
Related to the issue of unnatural canonical utterances, another challenge is the presence of unlikely examples with convoluted logical forms that rarely appear in real data.
As discussed earlier in \cref{sec:exp}, $\Dcan$ contains around $30\%$ such unlikely canonical examples (\eg $\utt_4$). %Since these utterances have convoluted logic or do not resemble natural English sentences, their paraphrases are much noisier (\eg~$\utt'_{4,*}$).
Similar to the case of unnatural utterances, paraphrases of those logically unlikely examples are also much noisier (\eg~$\utt'_{4,*}$).
Empirically, we observe the paraphraser's accuracy is only around $30\%$ for utterances of such unlikely samples, compared to $70\%$ for the likely ones. 
%Empirically, we observe for unlikely samples, the accuracy of the paraphraser is only around $30\%$, while for 
The filtering model is also less effective on unlikely examples (false positives \filterfp).
These noisy samples will eventually hurt performance of the parser.
We leave modeling utterance naturalness as important future work.
%These noisy paraphrases could potentially hurt performance, especially for zero-shot methods without prior knowledge about utterance naturalness.

% While the naturalness-driven data selection method (\cref{sec:bridge_gaps:data_selection}) could filter most unlikely samples, 

\paragraph{Cost of Grammar Engineering}
Our approach relies on an expressive SCFG to bridge the language and logical gaps between synthetic and real data.
While in \cref{sec:bridge_gaps:idiomatic_rules} we have identified a set of representative categories of grammar patterns necessary to capture domain-specific language style, and attempted to standardize the process of grammar construction by designing idiomatic productions following those categories, grammar engineering still remains a non-trivial task. 
%First, 
One need to have a good sense of the idiomatic language patterns that would frequently appear in real-world data, which requires performing user study or access to sampled data.
Additionally, encoding those language patterns as production rules assumes that the user understands the grammar formalism ($\lambda$-calculus) used by our system, which could limit the applicability of the approach to general users.
Still, as discussed in \cref{sec:bridge_gaps:idiomatic_rules}, we remark that for users proficient in the grammar formalism, curating a handful of idiomatic production rules is still more efficient than labeling parallel samples to exhaustively cover compositional logical patterns and diverse language style, and the size of annotated samples required could be orders-of-magnitude larger compared to the size of the grammar. 
%compared to manually labeling large amount of parallel samples with diverse logical compositionality and language style, which is necessary to train a data-hungry neural parser.
% In line with \citet{Xu2020AutoQA}, our model automatically synthesizes complex compositional samples from user-specified productions, which are further paraphrased to diversify their language style.
%Moreover, the requirement of proficiency in SCFGs could be potentially mitigated 
Meanwhile, the process of creating production rules could potentially be simplified by allowing users to define them using natural language instead of $\lambda$-calculus logical rules, similar in the spirit of the studies on naturalizing programs using canonical language \citep{Wang2017NaturalizingAP,Shin2021ConstrainedLM,Herzig2021UnlockingCG}.

\vspace{-0.5em}
\section{Related Work}
\vspace{-0.5em}
\label{sec:related_work}

%The cost of data acquisition hurdles scaling semantic parsers to emerging domains where labeled data is scarce,
%and becomes even more problematic with the dominance of data-hungry neural semantic parsers~(\newcite{DBLP:conf/acl/DongL16,dong18coarsefine,yin18tranx,Wang2019RATSQLRS}, \textit{inter alia}).
To mitigate the paucity of labeled data, the field has explored various supervision signals.
Specifically, \textbf{weakly-supervised} methods leverage the denotations 
%(\ie~execution results) 
of utterances as indirect supervision~\citep{DBLP:conf/conll/ClarkeGCR10,krishnamurthy-mitchell-2012-weakly}, with programs modeled as latent variables~\citep{berant2013freebase,pasupat2015compositional}. Optimization is challenging due to the noisy binary reward of execution correctness~\citep{Agarwal2019LearningTG}, calling for better learning objectives~\citep{DBLP:conf/acl/GuuPLL17,wang-etal-2021-learning-executions} or efficient search algorithms for latent programs~\citep{krishnamurthy17constraint,DBLP:journals/corr/LiangBLFL16,liang18mapo,Muhlgay2019ValuebasedSI}.
%Next, \textbf{semi-supervised} learning models use both utterances annotated with programs and extra unlabeled ones, 
Next, \textbf{semi-supervised} models leverage extra unlabeled utterances, using techniques like self-training~\citep{konstas2017neural} or generative models~\cite{Kocisky2016,yin18acl}.
As a step further, \textbf{unsupervised} methods only use unlabeled utterances~\citep{cao2019duallearning}, and leverage  linguistic scaffolds (\eg~dependency trees) to infer programs with similar structures~\citep{poon13grounded}. 
Like our model, such methods use lexicons to capture alignments between NL phrases and logical predicates~\citep{Goldwasser2011ConfidenceDU}, while our method does not require real utterances.
Finally, methods based on \overnight/~\citep{wang15overnight} synthesize parallel corpora from SCFGs~\citep{Cheng2019LearningAE,xu2020schema2qa} or neural sequence models~\citep{Guo2018QuestionGF}, and attempt to bridge the gaps between canonical and real utterances via paraphrase detection~\citep{herzig19dontdetect} and generation~\citep{su17crossdomain,Shin2021ConstrainedLM}, or representation learning \citep{Marzoev2020UnnaturalLP}.
%Finally, methods based on \overnight/ automatically construct parallel corpora with synthesized examples~\citep{wang15overnight}, where programs and utterances are generated compositionally using SCFGs \citep{Cheng2019LearningAE,xu2020schema2qa} or neural sequence transduction models \citep{Guo2018QuestionGF}.
%The field also explored bridging the canonical and real utterances using paraphrase detection~\citep{herzig19dontdetect} and generation~\citep{su17crossdomain,Shin2021ConstrainedLM}, or representation learning \citep{Marzoev2020UnnaturalLP}.
%Our work advances this line using paraphrasers powered by pre-trained LMs~\citep{Xu2020AutoQA} and idiomatic productions to capture domain-specific language styles. 

\vspace*{-0.3em}
\section{Conclusion}
\vspace*{-0.3em}

%In this paper, we propose a zero-shot semantic parser that closes the language and logical gaps between synthetic and real data. Our model achieves competitive results on \scholar/ and \geo/.
In this paper, we propose a zero-shot semantic parser that closes the language and logical gaps between synthetic and real data. on \scholar/ and \geo/, our system outperforms other annotation-efficient approaches with zero labeled data.
%, our model outperforms other annotation-efficient approaches with zero labeled data.

\bibliography{main}

\begin{thebibliography}{54}
\expandafter\ifx\csname natexlab\endcsname\relax\def\natexlab#1{#1}\fi

\bibitem[{Agarwal et~al.(2019)Agarwal, Liang, Schuurmans, and
  Norouzi}]{Agarwal2019LearningTG}
Rishabh Agarwal, Chen Liang, Dale Schuurmans, and Mohammad Norouzi. 2019.
\newblock Learning to generalize from sparse and underspecified rewards.
\newblock In \emph{ICML}.

\bibitem[{Aky{\"u}rek et~al.(2021)Aky{\"u}rek, Akyurek, and
  Andreas}]{Akyrek2020LearningTR}
Ekin Aky{\"u}rek, Afra~Feyza Akyurek, and Jacob Andreas. 2021.
\newblock Learning to recombine and resample data for compositional
  generalization.
\newblock In \emph{Proceedings of ICLR}.

\bibitem[{Andreas et~al.(2020)Andreas, Bufe, Burkett, Chen, Clausman, Crawford,
  Crim, DeLoach, Dorner, Eisner, Fang, Guo, Hall, Hayes, Hill, Ho, Iwaszuk,
  Jha, Klein, Krishnamurthy, Lanman, Liang, Lin, Lintsbakh, McGovern,
  Nisnevich, Pauls, Petters, Read, Roth, Roy, Rusak, Short, Slomin, Snyder,
  Striplin, Su, Tellman, Thomson, Vorobev, Witoszko, Wolfe, Wray, Zhang, and
  Zotov}]{machines2020task-oriented}
Jacob Andreas, John Bufe, David Burkett, Charles Chen, Josh Clausman, Jean
  Crawford, Kate Crim, Jordan DeLoach, Leah Dorner, Jason Eisner, Hao Fang,
  Alan Guo, David Hall, Kristin Hayes, Kellie Hill, Diana Ho, Wendy Iwaszuk,
  Smriti Jha, Dan Klein, Jayant Krishnamurthy, Theo Lanman, Percy Liang,
  Christopher~H Lin, Ilya Lintsbakh, Andy McGovern, Aleksandr Nisnevich, Adam
  Pauls, Dmitrij Petters, Brent Read, Dan Roth, Subhro Roy, Jesse Rusak, Beth
  Short, Div Slomin, Ben Snyder, Stephon Striplin, Yu~Su, Zachary Tellman, Sam
  Thomson, Andrei Vorobev, Izabela Witoszko, Jason Wolfe, Abby Wray, Yuchen
  Zhang, and Alexander Zotov. 2020.
\newblock Task-oriented dialogue as dataflow synthesis.
\newblock \emph{Transactions of the Association for Computational Linguistics},
  8.

\bibitem[{Artzi and Zettlemoyer(2013)}]{artzi-zettlemoyer:2013:TACL}
Yoav Artzi and Luke Zettlemoyer. 2013.
\newblock Weakly supervised learning of semantic parsers for mapping
  instructions to actions.
\newblock \emph{Transaction of ACL}.

\bibitem[{Berant et~al.(2013)Berant, Chou, Frostig, and
  Liang}]{berant2013freebase}
Jonathan Berant, Andrew Chou, Roy Frostig, and Percy Liang. 2013.
\newblock Semantic parsing on freebase from question-answer pairs.
\newblock In \emph{Proceedings of EMNLP}.

\bibitem[{Cao et~al.(2019)Cao, Zhu, Liu, Li, and Yu}]{cao2019duallearning}
Ruisheng Cao, Su~Zhu, Chen Liu, Jieyu Li, and Kai Yu. 2019.
\newblock Semantic parsing with dual learning.
\newblock In \emph{Proceedings of ACL}.

\bibitem[{Cheng et~al.(2019)Cheng, Reddy, Saraswat, and
  Lapata}]{Cheng2019LearningAE}
Jianpeng Cheng, Siva Reddy, V.~Saraswat, and Mirella Lapata. 2019.
\newblock Learning an executable neural semantic parser.
\newblock \emph{Computational Linguistics}, 45:59--94.

\bibitem[{Clarke et~al.(2010)Clarke, Goldwasser, Chang, and
  Roth}]{DBLP:conf/conll/ClarkeGCR10}
James Clarke, Dan Goldwasser, Ming{-}Wei Chang, and Dan Roth. 2010.
\newblock Driving semantic parsing from the world's response.
\newblock In \emph{Proceedings of CoNLL}.

\bibitem[{Dasigi et~al.(2019)Dasigi, Gardner, Murty, Zettlemoyer, and
  Hovy}]{Dasigi2019IterativeSF}
Pradeep Dasigi, Matt Gardner, Shikhar Murty, Luke~S. Zettlemoyer, and Eduard~H.
  Hovy. 2019.
\newblock Iterative search for weakly supervised semantic parsing.
\newblock In \emph{Proceedings of NAACL-HLT}.

\bibitem[{Devlin et~al.(2019)Devlin, Chang, Lee, and
  Toutanova}]{Devlin2019BERTPO}
Jacob Devlin, Ming-Wei Chang, Kenton Lee, and Kristina Toutanova. 2019.
\newblock Bert: Pre-training of deep bidirectional transformers for language
  understanding.
\newblock In \emph{Proceedings of NAACL-HLT}.

\bibitem[{Dong and Lapata(2016)}]{DBLP:conf/acl/DongL16}
Li~Dong and Mirella Lapata. 2016.
\newblock Language to logical form with neural attention.
\newblock In \emph{Proceedings of ACL}.

\bibitem[{Finegan-Dollak et~al.(2018)Finegan-Dollak, Kummerfeld, Zhang,
  Ramanathan, Sadasivam, Zhang, and Radev}]{finegan18improving}
Catherine Finegan-Dollak, Jonathan~K. Kummerfeld, Li~Zhang, Karthik Ramanathan,
  Sesh Sadasivam, Rui Zhang, and Dragomir Radev. 2018.
\newblock Improving text-to-{SQL} evaluation methodology.
\newblock In \emph{Proceedings of ACL}.

\bibitem[{Fried et~al.(2018)Fried, Hu, Cirik, Rohrbach, Andreas, Morency,
  Berg-Kirkpatrick, Saenko, Klein, and Darrell}]{fried2018speaker}
Daniel Fried, Ronghang Hu, Volkan Cirik, Anna Rohrbach, Jacob Andreas,
  Louis-Philippe Morency, Taylor Berg-Kirkpatrick, Kate Saenko, Dan Klein, and
  Trevor Darrell. 2018.
\newblock Speaker-follower models for vision-and-language navigation.
\newblock In \emph{Proceedings of NeurIPS}.

\bibitem[{Goldwasser et~al.(2011)Goldwasser, Reichart, Clarke, and
  Roth}]{Goldwasser2011ConfidenceDU}
Dan Goldwasser, Roi Reichart, J.~Clarke, and D.~Roth. 2011.
\newblock Confidence driven unsupervised semantic parsing.
\newblock In \emph{Proceedings of ACL}.

\bibitem[{Guo et~al.(2018)Guo, Sun, Tang, Duan, Yin, Chi, Cao, Chen, and
  Zhou}]{Guo2018QuestionGF}
Daya Guo, Yibo Sun, Duyu Tang, Nan Duan, Jian Yin, Hong Chi, James Cao, Peng
  Chen, and M.~Zhou. 2018.
\newblock Question generation from sql queries improves neural semantic
  parsing.
\newblock In \emph{Proceedings of EMNLP}.

\bibitem[{Gupta et~al.(2018)Gupta, Shah, Mohit, Kumar, and Lewis}]{gupta18task}
Sonal Gupta, Rushin Shah, Mrinal Mohit, Anuj Kumar, and Mike Lewis. 2018.
\newblock Semantic parsing for task oriented dialog using hierarchical
  representations.
\newblock In \emph{Proceedings of EMNLP}.

\bibitem[{Guu et~al.(2017)Guu, Pasupat, Liu, and
  Liang}]{DBLP:conf/acl/GuuPLL17}
Kelvin Guu, Panupong Pasupat, Evan~Zheran Liu, and Percy Liang. 2017.
\newblock From language to programs: Bridging reinforcement learning and
  maximum marginal likelihood.
\newblock In \emph{Proceedings of {ACL}}.

\bibitem[{Herzig and Berant(2019)}]{herzig19dontdetect}
Jonathan Herzig and Jonathan Berant. 2019.
\newblock Don't paraphrase, detect! rapid and effective data collection for
  semantic parsing.
\newblock In \emph{Proceedings of EMNLP}.

\bibitem[{Herzig et~al.(2021)Herzig, Shaw, Chang, Guu, Pasupat, and
  Zhang}]{Herzig2021UnlockingCG}
Jonathan Herzig, Peter Shaw, Ming-Wei Chang, Kelvin Guu, Panupong Pasupat, and
  Yuan Zhang. 2021.
\newblock Unlocking compositional generalization in pre-trained models using
  intermediate representations.
\newblock \emph{ArXiv}, abs/2104.07478.

\bibitem[{Iyer et~al.(2017)Iyer, Konstas, Cheung, Krishnamurthy, and
  Zettlemoyer}]{iyer17user}
Srinivasan Iyer, Ioannis Konstas, Alvin Cheung, Jayant Krishnamurthy, and Luke
  Zettlemoyer. 2017.
\newblock Learning a neural semantic parser from user feedback.
\newblock In \emph{Proceedings of ACL}.

\bibitem[{Jia and Liang(2016)}]{Jia2016}
Robin Jia and Percy Liang. 2016.
\newblock Data recombination for neural semantic parsing.
\newblock In \emph{Proceedings of ACL}.

\bibitem[{Kocisk{\'{y}} et~al.(2016)Kocisk{\'{y}}, Melis, Grefenstette, Dyer,
  Ling, Blunsom, and Hermann}]{Kocisky2016}
Tom{\'{a}}s Kocisk{\'{y}}, G{\'{a}}bor Melis, Edward Grefenstette, Chris Dyer,
  Wang Ling, Phil Blunsom, and Karl~Moritz Hermann. 2016.
\newblock Semantic parsing with semi-supervised sequential autoencoders.
\newblock In \emph{Proceedings of EMNLP}.

\bibitem[{Konstas et~al.(2017)Konstas, Iyer, Yatskar, Choi, and
  Zettlemoyer}]{konstas2017neural}
Ioannis Konstas, Srinivasan Iyer, Mark Yatskar, Yejin Choi, and Luke
  Zettlemoyer. 2017.
\newblock Neural amr: Sequence-to-sequence models for parsing and generation.
\newblock In \emph{Proceedings of ACL}.

\bibitem[{Krishna et~al.(2020)Krishna, Wieting, and
  Iyyer}]{krishna-etal-2020-reformulating}
Kalpesh Krishna, John Wieting, and Mohit Iyyer. 2020.
\newblock Reformulating unsupervised style transfer as paraphrase generation.
\newblock In \emph{Proceedings of EMNLP}.

\bibitem[{Krishnamurthy et~al.(2017)Krishnamurthy, Dasigi, and
  Gardner}]{krishnamurthy17constraint}
Jayant Krishnamurthy, Pradeep Dasigi, and Matt Gardner. 2017.
\newblock Neural semantic parsing with type constraints for semi-structured
  tables.
\newblock In \emph{Proceedings of EMNLP}.

\bibitem[{Krishnamurthy and
  Mitchell(2012)}]{krishnamurthy-mitchell-2012-weakly}
Jayant Krishnamurthy and Tom Mitchell. 2012.
\newblock Weakly supervised training of semantic parsers.
\newblock In \emph{Proceedings of EMNLP}.

\bibitem[{Lake and Baroni(2018)}]{lake18generalization}
Brenden Lake and Marco Baroni. 2018.
\newblock Generalization without systematicity: On the compositional skills of
  sequence-to-sequence recurrent networks.
\newblock In \emph{Proceedings of ICML}.

\bibitem[{Lapata(2006)}]{lapata-2006-automatic}
Mirella Lapata. 2006.
\newblock \href {https://doi.org/10.1162/coli.2006.32.4.471} {Automatic
  evaluation of information ordering: Kendall{'}s tau}.
\newblock \emph{Computational Linguistics}, 32(4):471--484.

\bibitem[{Lewis et~al.(2020)Lewis, Liu, Goyal, Ghazvininejad, Mohamed, Levy,
  Stoyanov, and Zettlemoyer}]{lewis20bart}
M.~Lewis, Yinhan Liu, Naman Goyal, Marjan Ghazvininejad, A.~Mohamed, Omer Levy,
  Ves Stoyanov, and Luke Zettlemoyer. 2020.
\newblock {BART}: Denoising sequence-to-sequence pre-training for natural
  language generation, translation, and comprehension.
\newblock In \emph{Proceedings of ACL}.

\bibitem[{Li and Jagadish(2014)}]{Li2014ConstructingAI}
Fei Li and H.~Jagadish. 2014.
\newblock Constructing an interactive natural language interface for relational
  databases.
\newblock \emph{Proc. VLDB Endow.}, 8:73--84.

\bibitem[{Liang et~al.(2017)Liang, Berant, Le, Forbus, and
  Lao}]{DBLP:journals/corr/LiangBLFL16}
Chen Liang, Jonathan Berant, Quoc Le, Kenneth~D. Forbus, and Ni~Lao. 2017.
\newblock Neural symbolic machines: Learning semantic parsers on freebase with
  weak supervision.
\newblock In \emph{Proceedings of ACL}.

\bibitem[{Liang et~al.(2018)Liang, Norouzi, Berant, Le, and Lao}]{liang18mapo}
Chen Liang, Mohammad Norouzi, Jonathan Berant, Quoc~V Le, and Ni~Lao. 2018.
\newblock Memory augmented policy optimization for program synthesis and
  semantic parsing.
\newblock In \emph{Proceedings of NIPS}.

\bibitem[{Luong et~al.(2015)Luong, Pham, and Manning}]{luong2015effective}
Thang Luong, Hieu Pham, and Christopher~D. Manning. 2015.
\newblock Effective approaches to attention-based neural machine translation.
\newblock In \emph{Proceedings of EMNLP}.

\bibitem[{Marzoev et~al.(2020)Marzoev, Madden, Kaashoek, Cafarella, and
  Andreas}]{Marzoev2020UnnaturalLP}
Alana Marzoev, S.~Madden, M.~Kaashoek, Michael~J. Cafarella, and Jacob Andreas.
  2020.
\newblock Unnatural language processing: Bridging the gap between synthetic and
  natural language data.
\newblock \emph{ArXiv}, abs/2004.13645.

\bibitem[{Muhlgay et~al.(2019)Muhlgay, Herzig, and
  Berant}]{Muhlgay2019ValuebasedSI}
Dor Muhlgay, Jonathan Herzig, and Jonathan Berant. 2019.
\newblock Value-based search in execution space for mapping instructions to
  programs.
\newblock In \emph{Proceedings of NAACL}.

\bibitem[{Pasupat and Liang(2015)}]{pasupat2015compositional}
Panupong Pasupat and Percy Liang. 2015.
\newblock Compositional semantic parsing on semi-structured tables.
\newblock In \emph{Proceedings of ACL}.

\bibitem[{Poon(2013)}]{poon13grounded}
Hoifung Poon. 2013.
\newblock Grounded unsupervised semantic parsing.
\newblock In \emph{Proceedings of ACL}.

\bibitem[{Radford et~al.(2019)Radford, Wu, Child, Luan, Amodei, and
  Sutskever}]{Radford2019LanguageMA}
Alec Radford, Jeff Wu, R.~Child, David Luan, Dario Amodei, and Ilya Sutskever.
  2019.
\newblock Language models are unsupervised multitask learners.

\bibitem[{Rajpurkar et~al.(2016)Rajpurkar, Zhang, Lopyrev, and
  Liang}]{Rajpurkar2016SQuAD10}
Pranav Rajpurkar, Jian Zhang, Konstantin Lopyrev, and Percy Liang. 2016.
\newblock Squad: 100, 000+ questions for machine comprehension of text.
\newblock In \emph{Proceedings of EMNLP}.

\bibitem[{Shaw et~al.(2019)Shaw, Massey, Chen, Piccinno, and
  Altun}]{Shaw2019GeneratingLF}
Peter Shaw, Philip Massey, Angelica Chen, Francesco Piccinno, and Y.~Altun.
  2019.
\newblock Generating logical forms from graph representations of text and
  entities.
\newblock In \emph{Proceedings of ACL}.

\bibitem[{Shin et~al.(2021)Shin, Lin, Thomson, Chen, Roy, Platanios, Pauls,
  Klein, Eisner, and Durme}]{Shin2021ConstrainedLM}
Richard Shin, C.~H. Lin, Sam Thomson, Charles Chen, Subhro Roy,
  Emmanouil~Antonios Platanios, Adam Pauls, D.~Klein, J.~Eisner, and
  Benjamin~Van Durme. 2021.
\newblock Constrained language models yield few-shot semantic parsers.
\newblock \emph{ArXiv}, abs/2104.08768.

\bibitem[{Su and Yan(2017)}]{su17crossdomain}
Yu~Su and Xifeng Yan. 2017.
\newblock Cross-domain semantic parsing via paraphrasing.
\newblock In \emph{Proceedings of EMNLP}.

\bibitem[{Wang et~al.(2021{\natexlab{a}})Wang, Lapata, and
  Titov}]{wang-etal-2021-learning-executions}
Bailin Wang, Mirella Lapata, and Ivan Titov. 2021{\natexlab{a}}.
\newblock Learning from executions for semantic parsing.
\newblock In \emph{Proceedings of NAACL}.

\bibitem[{Wang et~al.(2021{\natexlab{b}})Wang, Yin, Lin, and
  Xiong}]{Wang2021LearningTS}
Bailin Wang, Wenpeng Yin, Xi~Victoria Lin, and Caiming Xiong.
  2021{\natexlab{b}}.
\newblock Learning to synthesize data for semantic parsing.
\newblock In \emph{Proceedings of NAACL}.

\bibitem[{Wang et~al.(2017)Wang, Ginn, Liang, and
  Manning}]{Wang2017NaturalizingAP}
Sida~I. Wang, Samuel Ginn, Percy Liang, and Christopher~D. Manning. 2017.
\newblock Naturalizing a programming language via interactive learning.
\newblock In \emph{ACL}.

\bibitem[{Wang et~al.(2015)Wang, Berant, and Liang}]{wang15overnight}
Yushi Wang, Jonathan Berant, and Percy Liang. 2015.
\newblock Building a semantic parser overnight.
\newblock In \emph{Proceedings of {ACL}}.

\bibitem[{Wieting and Gimpel(2018)}]{Wieting2018ParaNMT50MPT}
J.~Wieting and Kevin Gimpel. 2018.
\newblock Paranmt-50m: Pushing the limits of paraphrastic sentence embeddings
  with millions of machine translations.
\newblock In \emph{Proceedings of ACL}.

\bibitem[{Xu et~al.(2020{\natexlab{a}})Xu, Campagna, Li, and
  Lam}]{xu2020schema2qa}
Silei Xu, Giovanni Campagna, Jian Li, and Monica~S Lam. 2020{\natexlab{a}}.
\newblock Schema2qa: High-quality and low-cost q\&a agents for the structured
  web.
\newblock In \emph{Proceedings of CIKM}.

\bibitem[{Xu et~al.(2020{\natexlab{b}})Xu, Semnani, Campagna, and
  Lam}]{Xu2020AutoQA}
Silei Xu, Sina~J. Semnani, Giovanni Campagna, and M.~Lam. 2020{\natexlab{b}}.
\newblock Autoqa: From databases to qa semantic parsers with only synthetic
  training data.
\newblock In \emph{Proceedings of EMNLP}.

\bibitem[{Yin and Neubig(2017)}]{yin17acl}
Pengcheng Yin and Graham Neubig. 2017.
\newblock \hypertarget{YN17}{}{A} syntactic neural model for general-purpose
  code generation.
\newblock In \emph{Proceedings of {ACL}}.

\bibitem[{Yin et~al.(2018)Yin, Zhou, He, and Neubig}]{yin18acl}
Pengcheng Yin, Chunting Zhou, Junxian He, and Graham Neubig. 2018.
\newblock Struct{VAE}: Tree-structured latent variable models for
  semi-supervised semantic parsing.
\newblock In \emph{Proceedings of ACL}.

\bibitem[{Yu et~al.(2018)Yu, Zhang, Yang, Yasunaga, Wang, Li, Ma, Li, Yao,
  Roman, Zhang, and Radev}]{Yu2018SpiderAL}
Tao Yu, Rui Zhang, Kai Yang, Michihiro Yasunaga, Dongxu Wang, Zifan Li, James
  Ma, Irene Li, Qingning Yao, Shanelle Roman, Zilin Zhang, and Dragomir~R.
  Radev. 2018.
\newblock Spider: A large-scale human-labeled dataset for complex and
  cross-domain semantic parsing and text-to-sql task.
\newblock In \emph{Proceedings of EMNLP}.

\bibitem[{Zelle and Mooney(1996)}]{geoquery}
John~M. Zelle and Raymond~J. Mooney. 1996.
\newblock Learning to parse database queries using inductive logic programming.
\newblock In \emph{Proceedings of AAAI}.

\bibitem[{Zhang et~al.(2019)Zhang, Baldridge, and He}]{paws2019naacl}
Yuan Zhang, Jason Baldridge, and Luheng He. 2019.
\newblock {PAWS: Paraphrase Adversaries from Word Scrambling}.
\newblock In \emph{NAACL}.

\end{thebibliography}
\bibliographystyle{acl_natbib}

\clearpage
\newpage
\onecolumn
\appendix

\begin{center}
\Large
\textbf{On The Ingredients of an Effective Zero-shot Semantic Parser}

Supplementary Materials
\end{center}

\section{Paraphraser}
\label{app:paraphraser}

Central to our approach is a paraphrase generation model $p(\utt \mapsto \utt')$, which paraphrases a canonical utterance $\utt$ to an alternative sentence $\utt'$ that is possibly more natural and linguistically diverse.
To improve the diversity of generated paraphrases, we paraphrase $\utt$ to multiple candidate rewrites $\{ \utt' \}$ using beam search.
We tested multiple strategies (\eg~nucleus sampling) to improve diversity of paraphrases via ensuring quality, and found beam search yields the best end-to-end performance.
%Each paraphrase $\utt'$ is then paired with the original program of $\utt$ to create a new example.

To generate high-quality paraphrases for open-domain utterances, we parameterize $p(\utt \mapsto \utt')$ using generative pre-trained LMs (\textsc{Bart}$_\textrm{Large}$).\footnote{We use the official implementation in {\tt fairseq}, \url{https://github.com/pytorch/fairseq}.}
The LM is fine-tuned on a corpus of $70K$ high-quality paraphrases sub-sampled from \textsc{ParaNmt}~\citep{Wieting2018ParaNMT50MPT} released by \citet{krishna-etal-2020-reformulating}, where samples are carefully constructed to ensure the lexical and syntactical diversity of target paraphrases.
To further improve the syntactic diversity of paraphrases from statement-style inputs (\eg~$\utt_2$, \cref{fig:system}), we apply force decoding with WH-prefixes (\eg~\textit{What}, \textit{When}, \textit{How many}, based on the answer type) to half hypotheses in the beam to generate question paraphrases (\eg~paraphrases prefixed with \textit{``How many''} for $\utt_3$ in \cref{fig:system}).

\paragraph{Filtering Paraphrases} 
While our paraphraser is strong, it is still far from perfect, especially when tasked with paraphrasing utterances found in arbitrary down-stream domains. 
For example, two ambiguous utterances ``\utterance{Author that cites A}'' and ``\utterance{Author cited by A}'' could get the same paraphrase ``\utterance{Who cites A?}''. Such noisy paraphrases will bring noise to learning and hurt performance.
To filter potentially incorrect outputs, we follow \citet{Xu2020AutoQA} and use the parser trained on the paraphrased data generated in the preceding iteration (or the seed canonical data at the beginning of training) to parse each paraphrased utterance, and only retain those for which the parser could successfully predict its program.
Admittedly such a stringent criterion will sacrifice recall, but empirically we found it works well. We present more analysis in the case study in \cref{sec:exp}.
%, outperforming more sophisticated approaches like filtering using strong paraphrase identification models. We present more insights in \cref{sec:exp}.

\section{Synchronous Grammar}
\label{app:grammar}

Our synchronous grammar is adapted from \citet{herzig19dontdetect} and \citet{wang15overnight}, which specifies alignments between NL expressions and logical form constituents in $\lambda$-calculus s-expressions.\footnote{We use the implementation in {\tt Sempre}, \url{https://github.com/percyliang/sempre}}
The grammar consists of a set of domain-general production rules, plus domain-specific rules specifying lexicons and idiomatic productions.
Specifically, domain-general productions define (1) generic logical operations like {\tt count} and {\tt superlative} (\eg~$r_3$, \cref{fig:system}), and (2) compositional rules to construct utterances following English syntax  (\eg~$r_1$, \cref{fig:system}). 
Domain-specific rules, on the other hand, are typically used to define task-dependent lexicons like types (\eg~{\tt author}), entities (\eg~{\tt allen\_turing}),  and relations (\eg~{\tt citations}) in the database. This work also introduces idiomatic productions to specific common NL expression catered to a domain, as detailed later.

\begin{table}[t]
  \centering
  \small
  \begin{tabular}{llp{5.5cm}}
  \toprule
  \textbf{Id} & \textbf{Productions (Syntactic Body and Semantic Function)} & \textbf{Description} \\ \hline

  $r_1$ & {\tt NP}$\mapsto${\tt SuperlativeAdj} {\tt NP} & \textit{e.g.~most recent \fbox{?}} \\
        & \begin{lstlisting}[style=program]
lambda rel, sub (
  call superlative (var sub) (string max) (var rel))
\end{lstlisting} & lambda function to get the subject {\tt sub} with the largest relation {\tt rel} \\ \hline

  $r_2$ & {\tt NP}$\mapsto${\tt NP+CP} & A noun phrase head {\tt NP} and a complementary phrase body {\tt CP} (\textit{e.g.~\underline{paper} \underline{in deep learning}}) \\
        & \begin{lstlisting}[style=program]
IdentityFn
\end{lstlisting} & An identity function returning child program \\ \hline
  
  $r_3$ & {\tt NP+CP}$\mapsto${\tt UnaryNP} {\tt CP} & \textit{e.g.~paper in deep learning} \\
        & \begin{lstlisting}[style=program]
Lambda Beta Reduction: f(var x)
\end{lstlisting} & Perform beta reduction, applying the function from {\tt CP} (\textit{e.g.~in deep learning}) to the value of {\tt UnaryNP} (\textit{e.g.~paper}) \\ \hline

  $r_4$ & {\tt UnaryNP}$\mapsto${\tt TypeNP} {\tt CP} & \multirow{2}{=}{Entity types, \eg~\textit{paper}} \\
        & \begin{lstlisting}[style=program]
IdentityFn
\end{lstlisting} & \\ \hline

  $r_5$ & {\tt CP}$\mapsto${\tt FilterCP} & \multirow{2}{=}{---} \\
        & \begin{lstlisting}[style=program]
IdentityFn
\end{lstlisting} & \\ \hline

  $r_6$ & {\tt FilterCP}$\mapsto${\tt Prep} {\tt NP} & \textit{e.g.~in deep learning} \\
        & \begin{lstlisting}[style=program]
lambda rel, obj, sub (
  call filter (var sub) (var rel) (string =) (var obj))
\end{lstlisting} & Create a lambda function, which filters entities in a list {\tt sub} such that its relation {\tt rel} (\textit{e.g.~topic}) equals {\tt obj} (\textit{e.g.~deep learning}) \\ \hline

  $r_5$ & {\tt NP}$\mapsto${\tt Entity} & \multirow{2}{=}{Entity noun phrases \textit{e.g.~deep learning}} \\
        & \begin{lstlisting}[style=program]
IdentityFn
\end{lstlisting} & \\ \bottomrule

  \end{tabular}
  \caption{Example domain-general productions rules in the SCFG}
  \label{tab:app:scfg_rules}
\end{table}

\begin{figure}[t]
  \centering
  \begin{minipage}{0.45\textwidth}
    \centering
    \includegraphics[width=0.85\textwidth]{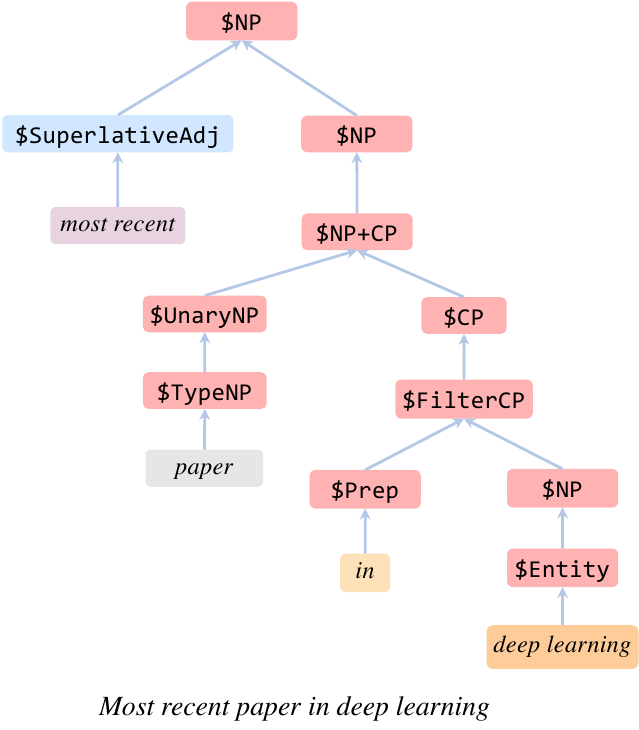}
  \end{minipage}\hfill
  \begin{minipage}{0.5\textwidth}
    \centering
    \begin{lstlisting}[style=program]
(
  call listValue (
      call superlative 
        (
          call filter 
          (
              call getProperty 
              (call singleton fb:en.paper) 
              (string ! type)
          ) 
          (string paper.keyphrase) 
          (string =) 
          fb:en.keyphrase.deep_learning
        ) 
      (string max) 
      (string paper.publication_year)
    )
)
\end{lstlisting}
\end{minipage}
\caption{(a) The derivation tree (production rule applications) to generate the example utterance and its program. (b) The program defined in s-expression.}
\label{fig:app:example_derivation}
\end{figure}

\cref{tab:app:scfg_rules} lists example domain-general productions in our SCFG.
\cref{fig:app:example_derivation} shows the derivation that applies those productions to generate an example utterance and program.
Each production has a syntactic body, specifying how lower-level syntactic constructs are composed to form more compositional utterances, as well as a \emph{semantic function}, which defines how programs of child nodes are composed to generate a new program.
For instance, the production $r_3$ in \cref{tab:app:scfg_rules} generates a noun phrase from a unary noun phrase {\tt UnaryNP} (\eg~\textit{paper}) and a complementary phrase {\tt CP} (\eg~\textit{in deep learning}) by concatenating the child nodes {\tt UnaryNP} and {\tt CP} (\eg~\textit{paper in deep learning}). On the program side, the programs of two child nodes on \cref{fig:app:example_derivation} are:

\newpage

\lstdefinestyle{inline}{%
  basicstyle=\fontfamily{cmtt}\small,
  columns=fullflexible,
  frame=bt,
  morecomment=[l][\color{red}]{\#},
  escapechar=\&
  }
\newcommand*{\Comment}[1]{\makebox[10.0cm][l]{\textcolor{red}{\# #1}}}%

\begin{lstlisting}[style=inline]
  &\Comment{Get all entities whose type is paper}&
  $UnaryNP: call getProperty (call singleton fb:en.paper) (string !type) 
  &\Comment{A lambda function that returns entities in x whose relation paper.keyphrase is deep\_learning}&
  $CP: lambda x (call 
             filter (x) 
             (string paper.keyphrase) 
             (string =) 
             (fb:en.keyphrase.deep_learning)
       )
\end{lstlisting}

Where the program of {\tt UnaryNP} is an entity set of papers, and the program of {\tt NP} is a lambda function with a variable {\tt x}, which filters the entity set. The semantic function of $r_3$ specifies how these two programs should be composed to form the program of their parent node {\tt NP$+$CP}, which performs $\beta$ reduction, assigning the entity set returned by {\tt UnaryNP} to the variable {\tt x}:
\begin{lstlisting}[style=inline]
  &\Comment{Get all papers whose keyphrase is deep learning}&
  $NP+CP: (call 
              filter (
                call getProperty (call singleton fb:en.paper) (string !type) 
              ) 
              (string paper.keyphrase) 
              (string =) 
              (fb:en.keyphrase.deep_learning)
           )
\end{lstlisting}

\begin{table}[t]
  \centering
  \small
  \begin{tabular}{llp{5cm}}
  \toprule
  Id & Production Body (Child Nodes and Semantic Function) & Description \\ \hline

  $r_1$ & {\tt RelVP}$\mapsto$\textit{publish in} & \multirow{2}{=}{Verb phrase for multi-hop relation \textit{author \underline{that writes paper in} ACL}} \\
        & \begin{lstlisting}[style=program]
ConstantFn (string author.publish_in)
\end{lstlisting} &  \\ \hline

  $r_2$ & {\tt SuperlativeAdj}$\mapsto$\textit{most recent} & \multirow{2}{=}{Superlative adjectives to describe publication dates} \\
        & \begin{lstlisting}[style=program]
ConstantFn (string paper.publication_year)
\end{lstlisting} &  \\ \hline

$r_3$ & {\tt SuperlativeMinAdj}$\mapsto$\textit{first} & \multirow{2}{=}{Superlative adjectives to describe the earliest publication dates} \\
        & \begin{lstlisting}[style=program]
ConstantFn (string paper.publication_year)
\end{lstlisting} &  \\ \hline

  $r_4$ & {\tt SuperlativeAdj}$\mapsto$\textit{published before} & \multirow{2}{=}{Comparative prepositions to describe publication dates} \\
        & \begin{lstlisting}[style=program]
ConstantFn (string paper.publication_year)
\end{lstlisting} &  \\ \hline

$r_5$ & {\tt CountSuperlativeNP}$\mapsto$\textit{the most popular topic for} & \multirow{2}{=}{ Superlative form to refer to the most frequent keyphrase for papers } \\
        & \begin{lstlisting}[style=program]
ConstantFn (string keyphrase.paper)
\end{lstlisting} &  \\ \hline

$r_6$ & {\tt MacroVP}$\mapsto$\textit{publish mostly in} & \multirow{2}{=}{ Superlative form of verb relational phrases with complex computation. {\tt countSuperlative} returns the entity {\tt x} in {\tt venue} for which the papers in {\tt x} (via relation {\tt venue.paper}) has the largest intersection with papers by {\tt author} (via realtion {\tt author.paper}) } \\
        & \begin{lstlisting}[style=program]
lambda author, venue (
  call countSuperlative 
    (var venue) 
    (string max) 
    (string venue.paper) 
    (call getProperty (var author) (string author.paper)) 
  )
\end{lstlisting} &  \\ \bottomrule

  \end{tabular}
  \caption{Example idiomatic productions used in \scholar/}
  \label{tab:app:idiomatic_prod:scholar}
\end{table}

\subsection{Idiomatic Productions}

\paragraph{Multi-hop Relations} We create idiomatic productions for non-compositional NL phrases of multi-hop relations (\eg~\utterance{Author that \underline{writes} paper \underline{in} ACL}). We augment the database with entries for those multi-hop relations (\eg~$\langle${\tt X}, {\tt author.publish\_in}, {\tt acl}$\rangle$), and then create productions in the grammar aligning those relations with their NL phrases (\eg~$r_1$ in \cref{tab:app:idiomatic_prod:scholar}). 

\paragraph{Comparatives and Superlatives} We also create productions for idiomatic comparatives and superlative expressions. 
Those productions specify the NL expressions for the comparative/superlative form of some relations. 
For example, for the relation {\tt paper.publication\_year} with objects of date time, its superlative form would be \textit{most recent} ($r_2$ in \cref{tab:app:idiomatic_prod:scholar}) and \textit{first} ($r_3$), while its comparative form could be prepositional phrases like \textit{published before} ($r_4$) and \textit{published after}. Those productions define the lexicons for comparative/superlative expressions, and could be used by the domain-general rules like $r_1$ in \cref{tab:app:scfg_rules} to compose utterances (\eg~\cref{fig:app:example_derivation}).

Besides superlative expressions for relations whose objects are measurable, we also create idiomatic expressions for relations with countable subjects or objects.
As an example, the utterance ``\utterance{\underline{The most popular topic for} papers in ACL}'' involves grouping ACL papers by topic and return the most frequent one. Such computation is captured by the {\tt CountSuperlative} operation in our SCFG based on \citet{wang15overnight}, and we create productions aligning those relations with the idiomatic noun phrases describing their superlative form (\eg~$r_5$ in \cref{tab:app:idiomatic_prod:scholar}).

Perhaps the most interesting form of superlative relations are those involving reasoning with additional entities. 
For instance, the relation in \textit{``venue that X \underline{publish mostly in}''} between the entity {\tt author} and {\tt venue} implicitly involves counting the {\tt paper}s that the author \textit{X} publishes.
%Intuitively we could treat these relations similarly as multi-hop ones, and create  
For those relations, we create ``macro'' productions (\eg~$r_6$ in \cref{tab:app:idiomatic_prod:scholar}), which defines the lambda function that computes the answer (\eg~return the publication venue where \textit{X} publishes the most number of papers) given the arguments (\eg~an author \textit{X}).

% %and their non-compositional surface phrases by treating those relations as single-hop ones.
% %We first add entries in the database capturing such relations (), and 
% %then create productions aligning the phrases to the new DB relations.
% % (\eg~xxx on Figure xxx). 

\section{Model Configurations}
\label{app:config}

\paragraph{Paraphraser} We finetune the paraphraser using a batch size of $1,024$ tokens for $5,000$ iterations ($500$ for warm-up), with a learning rate of $3e-5$ using \textsc{Adam}. We apply label smoothing with a probability of $0.1$. %Paraphrases are generated from the model using beam search of size 10.

\paragraph{Semantic Parser} Our semantic parser is a neural sequence-to-sequence model with dot-product attention~\citep{luong2015effective}, using a \textsc{Bert}$_\textrm{Base}$ encoder and an LSTM decoder, augmented with copying mechanism. The size of the LSTM hidden state is 256. We decode programs using beam search with a beam size of 5. Following \citet{herzig19dontdetect}, we remove hypotheses from the beam that leads to error executions.

\paragraph{Iterative Training} As described in \cref{sec:bridge_gaps:idiomatic_rules}, we first run the iterative paraphrasing and training algorithm for one pass to generate the validation set. 
In the first iteration of this stage, we train a semantic parser on the (unparaphrased) seed canonical data ($\Dcan$) as the initial paraphrase filtering model.
In the second stage, we restart the learning process using the generated validation set, and initialize the paraphrase filtering model using the previously trained semantic parser.
For each stage, we run the iterative learning algorithm (\cref{sec:system}) for two iterations. We generate 10 paraphrases for each example. In each iteration, we train the semantic parser for 30 epochs with a batch size of 64. We use separate learning rates for the \textsc{Bert} encoder ($3e-5$) and other parameters ($0.001$) in the model~\citep{Shaw2019GeneratingLF}. 
For each iteration in the second stage, we perform validation by finding the model checkpoint that achieves the lowest perplexity on the validation set.
We perform validation using perplexity for efficiency reasons, as evaluating denotation accuracy requires performing beam search decoding and querying the database, which could be slow.

\paragraph{Evaluation Metric} 
For the perplexity metric to evaluate language gaps, we fine-tune a \textsc{Gpt-2} language model on the paraphrased canonical data $\Dpara$ for $1,500$ steps ($150$ steps for warm-up) with a batch size of 64 and a learning rate of $1e-5$. 
We use the following equation to compute perplexity
\begin{equation}
\textrm{PPL}(\Dnat) = \exp \big( \frac{1}{| \Dnat |} \sum_{ \langle \utt, \mr \rangle \in \Dnat } \frac{ - \log p(\utt) } { | \utt | } \big)
\label{eq:ppl}
\end{equation}
This is slightly different from the standard corpus-level perplexity. We use this metric because it is more sensitive (larger $\Delta$) on our small ($<1K$) evaluation sets, and always correlates with the corpus-level perplexity. For reference, here is the sequence of perplexities using \cref{eq:ppl} in the upper half of \cref{exp:tab:scholar_geo_with_train_size_main} compared to the corpus-level ones:
\begin{table}[h!]
  \small
  \centering
  \begin{tabular}{lccccc}
  \cref{eq:ppl} & 22.0 & 21.4 & 20.9 & 20.7 & 21.5 \\ \hline
  Corpus-PPL    & 19.3 & 18.8 & 18.4 & 18.2 & 18.8  \\
  \end{tabular}
\end{table}

\end{document}